\documentclass[10pt,twocolumn,letterpaper]{article}

\usepackage{cvpr}
\usepackage{times}
\usepackage{epsfig}
\usepackage{graphicx}
\usepackage{amsmath}
\usepackage{amssymb}
\usepackage{subcaption}
\usepackage{multirow}
\usepackage{graphics}
\usepackage{pifont}
\usepackage{float}
\usepackage{scrtime}
\usepackage{authblk}

\usepackage[breaklinks=true,bookmarks=false]{hyperref}

\newcommand{\cmark}{\ding{51}}%
\newcommand{\xmark}{\ding{55}}%

\newcommand*{\affaddr}[1]{#1} 
\newcommand*{\affmark}[1][*]{\textsuperscript{#1}}
\newcommand*{\email}[1]{\small\texttt{#1}}

\cvprfinalcopy 

\def\cvprPaperID{****} 

\begin{document}

\title{Joint Learning of Portrait Intrinsic Decomposition and Relighting}

\author{%
Mona Zehni$^{*}$\affmark[1], Shaona Ghosh\affmark[2], Krishna Sridhar\affmark[2], Sethu Raman\affmark[2]\\
\affaddr{\affmark[1]Department of ECE and CSL, University of Illinois at Urbana-Champaign}\\
\affaddr{\affmark[2]Apple Inc.}\\
\email{mzehni2@illinois.edu, \{shaona\_ghosh,srikrishna\_sridhar,sethur\}@apple.com}
}

\maketitle
\footnotetext{
Mona Zehni was a summer 2020 intern at Apple while working on this project.}
\vspace{-20pt}
\begin{abstract}
\vspace{-10pt}
Inverse rendering is the problem of decomposing an image into its intrinsic components, i.e. albedo, normal and lighting. To solve this ill-posed problem from single image, state-of-the-art methods in shape from shading mostly resort to supervised training on all the components on either synthetic or real datasets. Here, we propose a new self-supervised training paradigm that 1) reduces the need for full supervision on the decomposition task and 2) takes into account the relighting task. We introduce new self-supervised loss terms that leverage the consistencies between multi-lit images (images of the same scene under different illuminations). Our approach is applicable to multi-lit datasets. We apply our training approach in two settings: 1) train on a mixture of synthetic and real data, 2) train on real datasets with limited supervision. We showcase the effectiveness of our training paradigm on both intrinsic decomposition and relighting and demonstrate how the model struggles in both tasks without the self-supervised loss terms in limited supervision settings. We provide results of comprehensive experiments on SfSNet, CelebA and Photoface datasets and verify the performance of our approach on images in the wild. 
\end{abstract}

\begin{figure}
    \centering
    \includegraphics[width=0.45\textwidth]{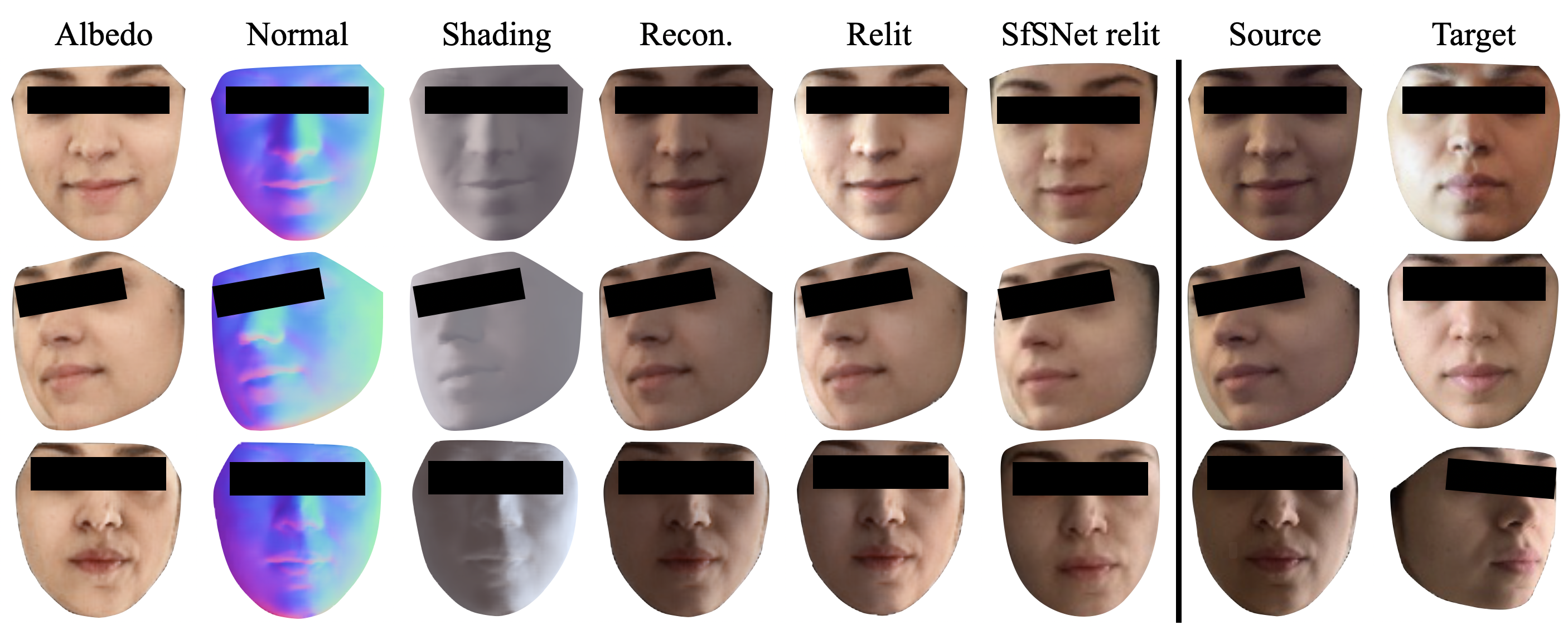}
    \captionof{figure}{Estimated shape and relighting from a single image. Images in the wild taken with a fourth generation iPad Pro  device. Our model is trained with light supervision and self-supervised loss terms without any supervision on albedo and normals or training on synthetic data. Re-lit images are obtained by transferring the estimated lighting from the target to the source image. We also compare our model against SfSNet~\cite{sfsnetSengupta18}. Partially redacted face of the subject is used with permission.}
    \label{fig:my_label}
\end{figure}
\vspace{-20pt}
\section{Introduction}
Image decomposition into its intrinsic components (albedo, normals and lighting), i.e. inverse rendering is a fundamental problem in computer vision~\cite{tappen2005, tang2012deep}. Not only it allows us to understand the geometry of the scene, it also enables us to manipulate different components and generate a desired output image. For example, by changing the illumination while fixing the albedo and the surface normals, a relit image of the same scene is obtained. Inverse rendering has a wide range of applications in image editing, subject relighting and augmented reality.

On a single image specifically, intrinsic decomposition is extremely challenging. Under Lambertian assumption, this originates from an underconstrained system of equations, thus leading to ambiguity and non-uniqueness of the decomposition solution. In such scenarios having strong priors on the individual components help in accurate image decomposition. Deep learning models that are trained on large datasets have proven to form strong priors in various ambiguous settings such as 3D reconstruction from single view image~\cite{Wu_2018_ECCV, choy2016}, image inpainting~\cite{Yeh_2017_CVPR} and super-resolution~\cite{Kim_2016_CVPR}, to name a few.

Here, we also resort to a deep learning based data-driven approach for the single-image intrinsic decomposition task. Previous state-of-the-art methods have tackled this problem through supervised training on all intrinsic components either on synthetic~\cite{li2018learning, Narihira_2015}, real datasets~\cite{Nestmeyer_2020_CVPR} or a combination of both~\cite{sfsnetSengupta18}. Here, we propose a new semi self-supervised training paradigm that helps in learning the intrinsic decomposition and relighting tasks in regimes where ground truth labels for intrinsic components are not available. Our approach leverages the images of the same scene lit under different illuminations, i.e. multi-lit images, and their inherent consistencies during training, while operating on single images at inference. For this purpose, we generate a multi-lit dataset from the synthetic SfSNet~\cite{sfsnetSengupta18} and real CelebA~\cite{liu2015faceattributes} datasets, called \textit{multi-lit CelebA}. Compared to self-supervised methods such as~\cite{yu19inverserendernet} applied to scene datasets, our method does not require multi-view image datasets and allows for hybrid training on synthetic and real datasets. We evaluate the effect of our training approach in two settings: 1) hybrid training on real and synthetic data, 2) training on real datasets with limited supervision. 

We restate our contributions as: 1) We introduce a new self-supervised \textit{cross-relighting} loss. Assuming a pair of multi-lit images, cross-relighting term enforces consistency between relit images obtained after swapping illumination vectors between the pair. To facilitate the use of this loss term, we generate a multi-lit dataset by relighting images from CelebA~\cite{liu2015faceattributes} with lighting vectors sampled from the synthetic SfSNet dataset~\cite{sfsnetSengupta18}. 2) Through extensive experiments and comparison with multiple baselines, we demonstrate the importance of the cross-relighting loss term in either hybrid training on real and synthetic datasets or training with limited supervision on real datasets. We show that without this loss term, learning the intrinsic decomposition and relighting tasks fail in limited supervision settings.

\begin{figure*}
\includegraphics[width=1 \linewidth]{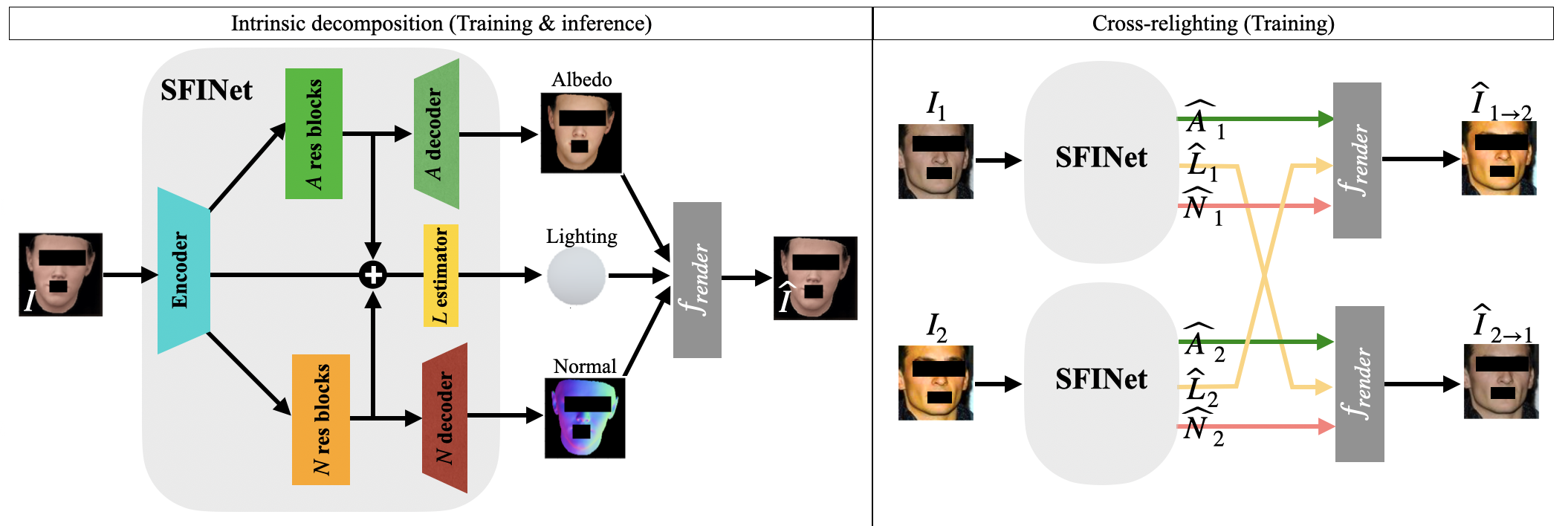}
\vspace{-10pt}
\caption{An illustration of our pipeline. SFINet (shape from image network) refers to the block which performs the intrinsic decomposition.}
\label{fig:arch}
\vspace{-10pt}
\end{figure*}

\section{Related Work}
Intrinsic decomposition and subject relighting have been extensively studied in the literature. Here, we summarize the existing methods.

\noindent \textbf{Intrinsic Image Decomposition}: Early works on intrinsic image decomposition are mainly derived by solving the equations provided by the Lambertian reflectance model~\cite{barrow1978recovering, land1971lightness, Zhang1999}. To impose proper constraints for intrinsic decomposition of single images which is a highly ill-posed problem, many papers introduced priors on different components~\cite{Malik2015}. However, deriving effective priors that model the underlying physics is challenging.

Recently, data driven models surged in the intrinsic decomposition field due to their exceptional performance in learning effective priors from large datasets. Deep belief networks~\cite{tang2012deep} and encoder-decoder architectures~\cite{kulkarni2015deep,NeuralFace2017,sfsnetSengupta18, li2018learning, Baslamisli_2018_ECCV} have particularly proven effective. In the latter, the image is encoded into a latent representation which is then processed by separate decoders, to achieve different components such as albedo, normals and lighting. These architectures are trained on 1) synthetic~\cite{Narihira_2015, li2018learning, Lettry2018, shi2017learning, Zhu2020}, 2) unconstrained real datasets~\cite{yu19inverserendernet}, or 3) a mix of both~\cite{sfsnetSengupta18, neuralSengupta19, chen2019photo, zhou2018label}. While models trained on purely synthetic data with full supervision obtain reasonable results on real images, they still suffer from the domain gap between the real and synthetic datasets. Training on real datasets addresses this domain gap. However, it introduces a new challenge: lack of ground truth labels for the intrinsic components. To resolve this issue, many papers pursue the generation of pseudo-labels for real portrait datasets by fitting a 3D morphable model (3DMM)~\cite{3DMM, NeuralFace2017, trigeorgis2017face}, using a pre-trained model on synthetic data~\cite{sfsnetSengupta18, neuralSengupta19}, or photometric stereo~\cite{Nestmeyer_2020_CVPR}. The main drawbacks of estimating these labels are: 1) The estimated components can be erroneous. This error then propagates to the trained models on these datasets, 2) Obtaining labels is often prohibitively expensive and not attainable for commercial use.

Self-supervised learning has gained increasing popularity to allow training on unlabeled datasets. For the intrinsic decomposition task, self-supervision is obtained by dense correspondence between pixels across multiple views~\cite{yu19inverserendernet}, training on a sequence of multi-lit images or video streams~\cite{BigTimeLi18, song2020recovering}, model-based shape reconstruction~\cite{mofa}, or through reconstruction loss (imposing consistency between the original images and the re-rendered one from the estimated intrinsic components) while training on a mix of labeled and unlabeled datasets~\cite{sfsnetSengupta18, neuralSengupta19, janner2017self}. Here, we introduce a new self-supervised loss term that 1) reduces the need for pseudo-labels and multi-stage training~\cite{sfsnetSengupta18}, 2) does not require a sequence of images as input during training~\cite{BigTimeLi18, song2020recovering}, 3) does not rely on strong priors posed in~\cite{yu19inverserendernet, BigTimeLi18, mofa} for training in limited supervision scenarios (no labels on albedos and normals exist) where the intrinsic decomposition from single image is highly ambiguous. Furthermore, compared to~\cite{ma2018single} proposing an unsupervised intrinsic decomposition technique given multi-lit images at training, we further disentangle the lighting component from the normals, thus facilitating relighting and light transfer between a source and a target image pair.

\noindent \textbf{Subject Relighting}: Relighting an image has been the focus of a multitude of recent works. \cite{xu2018, Meka2019} use multiple input images alongside the target lighting to generate the relit subject. While in~\cite{Meka2019}, these images are recorded under spherical color gradient illumination~\cite{Fyffe2009, Debevec2015}, in~\cite{xu2018} the input images correspond to multiple fixed directional lightings. These methods are in line with the idea that a new relit image is obtained by a combination of densely sampled images under one lighting at a time (OLAT) light stage~\cite{Debevec2000}. Encoder-decoder models estimating the source lighting and the image latent vector that is later on fed to the decoder alongside the target lighting to generate the relit image are proposed in~\cite{DPR, Sun2019}.  Furthermore, the task of light transfer from a source to a target image has been addressed as a mass transport problem in~\cite{Shu2017}. While these methods are capable of relighting subjects under different illuminations, they lack the interpretability that is put forward by the intrinsic decomposition-based relighting models. Nestmeyer et al in~\cite{Nestmeyer_2020_CVPR} proposed a physics-based model that relights an image given a target lighting while taking into account non-diffuse components. Joint relighting and multi-view synthesis from multiple input images has also been discussed in~\cite{xu2019deep, zhang2020neural, chen2020neural}. Opposed to our approach, these models rely on multiple input images at inference to extract the full geometry of the underlying scene.

\section{Method}
\label{sec:method}
Our goal is to decompose a single-view RGB image $I$ into its intrinsic components including albedo $A$, normal $N$ and lighting $L$. The albedo and normals have the same dimensionality as the image, i.e. $n \times n \times 3$. While for the albedo, different channels represent color, for normals, they mark the orientation of the normal along 3D axes. Furthermore, we represent colored lighting by its spherical harmonic (SH) coefficients up to second order, thus $L \in \mathbb{R}^{9 \times 3}$. While SH coefficients up to second order mainly constitute low-frequency terms, they are sufficient to estimate the irradiance up to $1\%$ error~\cite{Ramamoorthi2001}. We assume $I$ is generated under Lambertian reflectance as,
\begin{align}
I(p) = f_{\textrm{render}} (A(p), N(p), L) = A(p) \circ (L^T N_{\textrm{SH}}(p)) 
\label{eq:lambertian}
\end{align}
where $p$ marks the $p$-th pixel and $f_{\textrm{render}}$ is the differentiable function that maps the intrinsic components to the image domain. Also, $N_{\textrm{SH}}$ is the transformed normals in SH domain and $\circ$ denotes element-wise multiplication. 
\vspace{-3pt}
\subsection{Architecture}
\vspace{-3pt}
The decomposition architecture that we adopt here is inspired by the SfSNet architecture~\cite{sfsnetSengupta18} depicted in Fig.~\ref{fig:arch}, consisting of 1) a shared encoder, 2) two streams of cascaded residual blocks to extract albedo and normal features and 3) three decoders for albedo, normals and lighting. On the encoding path, the image is processed through convolution layers, batch norm~\cite{bn} and ReLU~\cite{relu} activations. On the decoding path, to upsample the downscaled features, we use deconvolution layers. To estimate the lighting, we concatenate the image, albedo and normal features and feed it to a trainable light estimator. This is inspired by the fact that given an image, its albedo and normals, lighting is estimated by solving a linear equation defined in~\eqref{eq:lambertian}. The light estimator block recovers the SH representation of lighting up to the second order.

\vspace{-3pt}
\subsection{Training}
\vspace{-3pt}
We adopt a combination of supervised and self-supervised strategies using a mix of real and synthetic datasets. Below, we describe our training methods specific to the different datasets and the available annotations. 

\noindent \textbf{Supervised}: For synthetic datasets, ground truth (GT) albedo, normal and lighting are available. Thus, these ground truth values are used to supervise the estimation of $A$, $N$ and $L$ during training. Further, through the reconstruction loss, we enforce the reconstructed image (combining the estimated components following~\eqref{eq:lambertian}) to be close to the original input image. 

\noindent \textbf{Self-supervised}: We also train our model on real datasets to improve generalization on unconstrained images in the wild. As ground truth intrinsic components are not often available for these datasets, we use self-supervision. Thus, apart from the reconstruction loss, we introduce a new term called \textit{cross-relighting} loss.

To explain cross-relighting consistency terms, let us consider two images $I_1$ and $I_2$. These images share the same albedo and normals and only differ in terms of lighting, where $I_1 \! = \! f_{\textrm{render}}(A_1, N_1, L_1)$ and $I_2 \! = \! f_{\textrm{render}}(A_2, N_2, L_2)$, with $A_1=A_2$ and $N_1=N_2$. Therefore, we expect the estimated albedos and normals for $I_1$ and $I_2$ to be the same. Furthermore, by swapping the estimated lighting for the two images, we expect $f_{\textrm{render}}(\widehat{A}_1, \widehat{N}_1, \widehat{L}_2)$ to be close to $I_2$ and $f_{\textrm{render}}(\widehat{A}_2, \widehat{N}_2, \widehat{L}_1)$ to be close to $I_1$. This adds further consistency between the relit images which has shown to be beneficial during training. Furthermore, we verify that the use of cross-relighting loss term reduces the need for providing pseudo-labels for the real unconstrained images~\cite{sfs}. Imposing the cross-relighting loss term is further illustrated in Fig.~\ref{fig:arch}.

Finally, the final loss that is used to train our architecture on a mix of synthetic and real datasets is defined as,
\vspace{-5pt}
{\small
\begin{align}
\mathcal{L} = & \lambda_{A} \Vert \widehat{A} \! - \!  A \Vert_1 \!  + \!  \lambda_{N} \Vert \widehat{N} \!  - N \! \Vert_1 \!  + \!  \lambda_{L} \Vert \widehat{L}\!  - \! L \Vert_2^2  \! +  \! \nonumber \\ 
& \lambda_{\textrm{rec}} \mathcal{L}_{\textrm{rec}} \!  + \!  \lambda_{\textrm{relit}} \mathcal{L}_{\textrm{relit}}
\label{eq:loss}
\vspace{-1pt}
\end{align}
}%
\noindent where the first three terms mark the supervision on albedo, normals and lighting and are only effective for samples from the synthetic dataset. Furthermore, $\mathcal{L}_{\textrm{rec}}$ and $\mathcal{L}_{\textrm{relit}}$ are the reconstruction and relighting consistency terms defined as,
{\small
\begin{align}
\mathcal{L}_{\textrm{rec}} \! &= d(f_{\textrm{render}}(\widehat{A}_1, \widehat{N}_1, \widehat{L}_1), I_1) + d(f_{\textrm{render}}(\widehat{A}_2, \widehat{N}_2, \widehat{L}_2), I_2) \nonumber \\
\mathcal{L}_{\textrm{relit}} \! &= d(f_{\textrm{render}}(\widehat{A}_1, \widehat{N}_1, \widehat{L}_2), I_2) + d(f_{\textrm{render}}(\widehat{A}_2, \widehat{N}_2, \widehat{L}_1), I_1) \nonumber
\end{align}}%
with $d(\widehat{x}, x)$ denoting the distance metric between $\widehat{x}$ and $x$. In our experiments, we choose $\Vert . \Vert_1$ as $d$ to better preserve the high frequency details in the reconstructed image and intrinsic components. We compute $\mathcal{L}_{\textrm{recon}}$ and $\mathcal{L}_{\textrm{relit}}$ in LAB color space, as it leads to a higher quality relighting results. 

Note that, we only need a multi-lit dataset for the training phase, while inference is performed on a single image. We use existing multi-lit dataset PhotoFace~\cite{photoface} and also generate a multi-lit dataset from single-lit CelebA~\cite{liu2015faceattributes} and synthetic SfSNet~\cite{sfsnetSengupta18} dataset. We call this dataset \textit{multi-lit CelebA}. Details are discussed in Section 4.1. 

\noindent \textbf{Non-hybrid training}: While we use~\eqref{eq:loss} to train on a mixture of synthetic and real multi-lit datasets, we also investigate a setting in which we only train on multi-lit datasets with lighting annotations. As discussed in Section 4.1, our multi-lit CelebA dataset consists of images of the same subjects under different lightings alongside the GT illumination vectors. In this setting, we train the same architecture in Fig.~\ref{fig:arch} with light supervision, $\mathcal{L}_{\textrm{recon}}$ and $\mathcal{L}_{\textrm{relit}}$ terms. Note that, here there is no longer any form of supervision on albedo and normals, as we are not training on synthetic data anymore. To provide some priors on the estimated albedos and normals, we make slight changes in the architecture. For the albedos, we add a sigmoid non-linearity to bound the estimated albedos between $[0, 1]$. Furthermore, to ensure that the normals have unit norms, we further normalize $N$ pixel-wise. In Section~\ref{sec:results}, we present results on this form of training and argue that without $\mathcal{L}_{\textrm{relit}}$, the model is not able to learn a meaningful decomposition between the albedos and the normals. This highlights the importance of $\mathcal{L}_{\textrm{relit}}$ in regimes with limited available annotations.

\begin{figure}
\centering
\begin{minipage}{0.55\linewidth}
\centering
\includegraphics[width=1\linewidth]{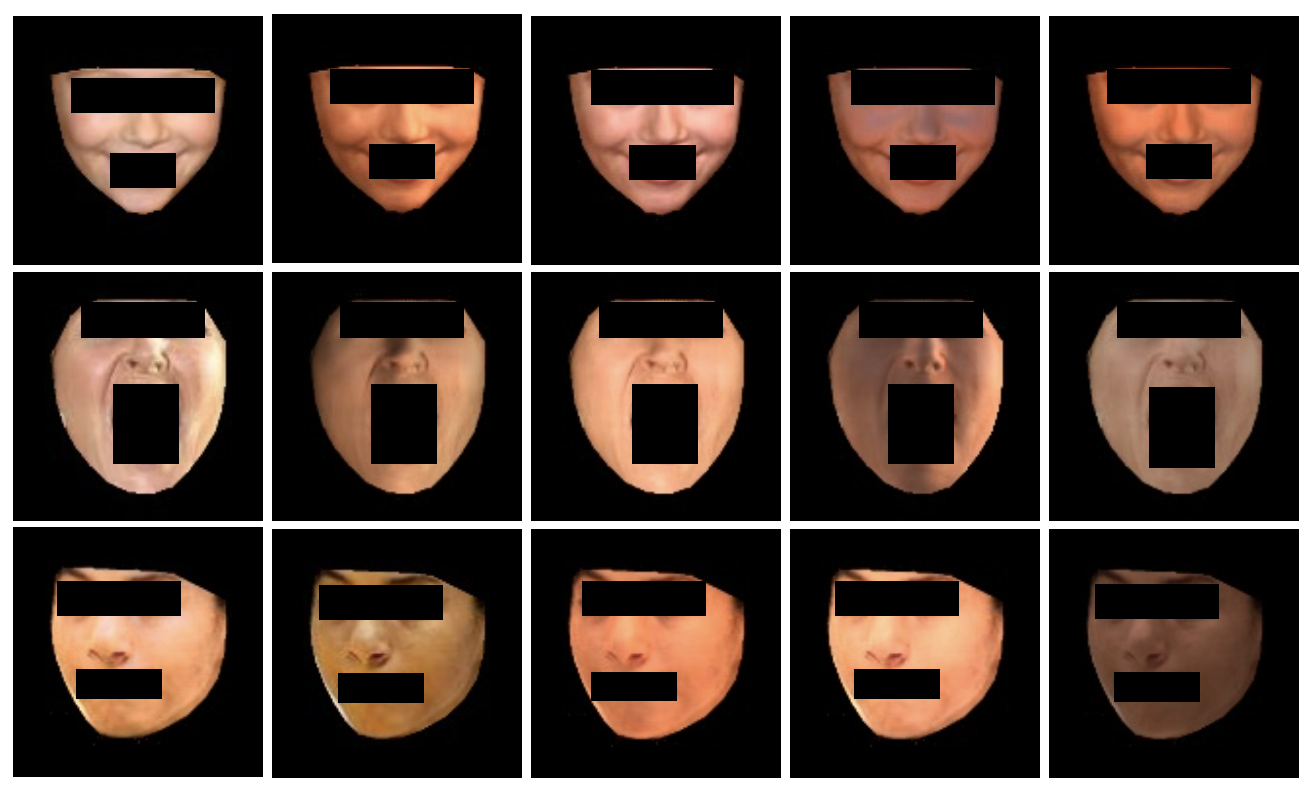}
\subcaption{Multi-lit CelebA}
\end{minipage}
\hfill
\begin{minipage}{0.4\linewidth}
\centering     
\includegraphics[width=0.9\linewidth]{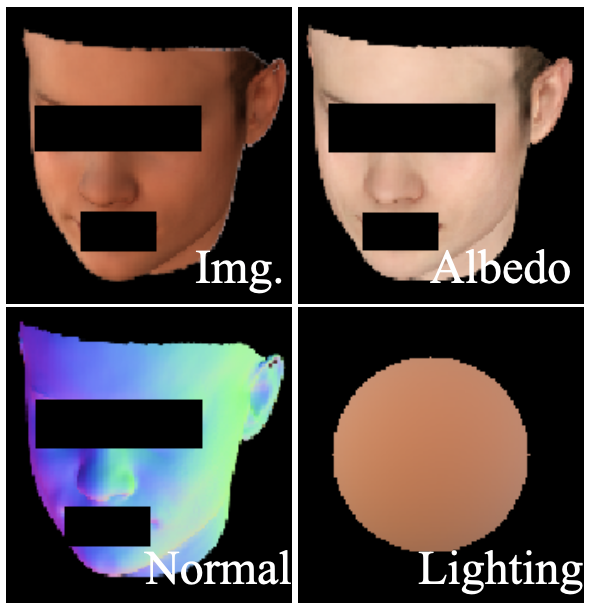}
\subcaption{SfSNet}
\end{minipage}
\vspace{-10pt}
\caption{Visualizing samples and annotations for (a) generated multi-lit CelebA~\cite{liu2015faceattributes}, 2) SfSNet~\cite{sfsnetSengupta18} dataset.}
\label{fig:dataset}
\vspace{-10pt}
\end{figure}

\section{Experiments}
\label{sec:results}

\begin{figure*}
\centering
\begin{minipage}{1\textwidth}
\centering
\includegraphics[width=1\linewidth]{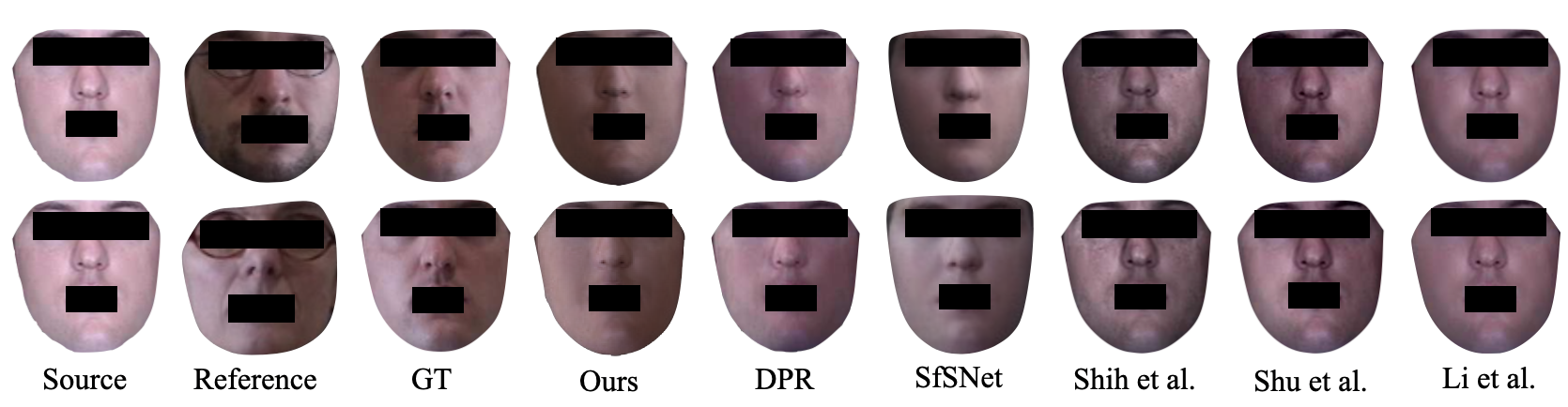}
\vspace{-10pt}
\subcaption{Light transfer on MultiPIE dataset~\cite{multipie}.}
\end{minipage}
\vspace{7pt}
\begin{minipage}{1\textwidth}
\centering     
\includegraphics[width=1\linewidth]{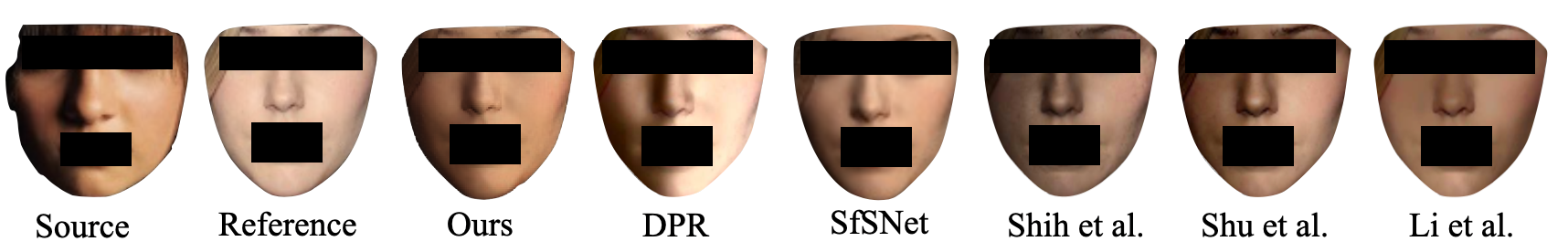}
\vspace{-10pt}
\subcaption{Light transfer on CelebA-HQ dataset~\cite{CelebAMask-HQ}.}
\end{minipage}
\vspace{7pt}
\begin{minipage}{1\textwidth}
\centering     
\includegraphics[width=1\linewidth]{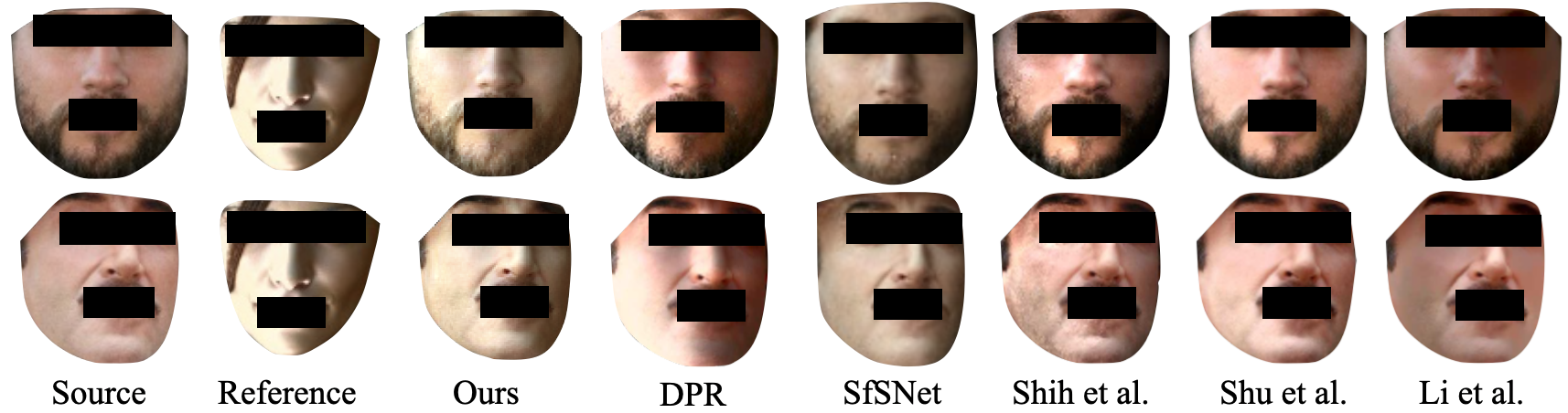}
\vspace{-10pt}
\subcaption{Light transfer from samples of DPR dataset~\cite{DPR} to CelebA-HQ dataset~\cite{CelebAMask-HQ}.}
\end{minipage}
\vspace{-10pt}
\caption{Relighting quality and comparison with multiple relighting baselines. We evaluated on the screenshot of the images provided by~\cite{DPR}.}
\label{fig:relight_baselines}
\vspace{-10pt}
\end{figure*}

\begin{figure}
\centering
\includegraphics[width=1\linewidth]{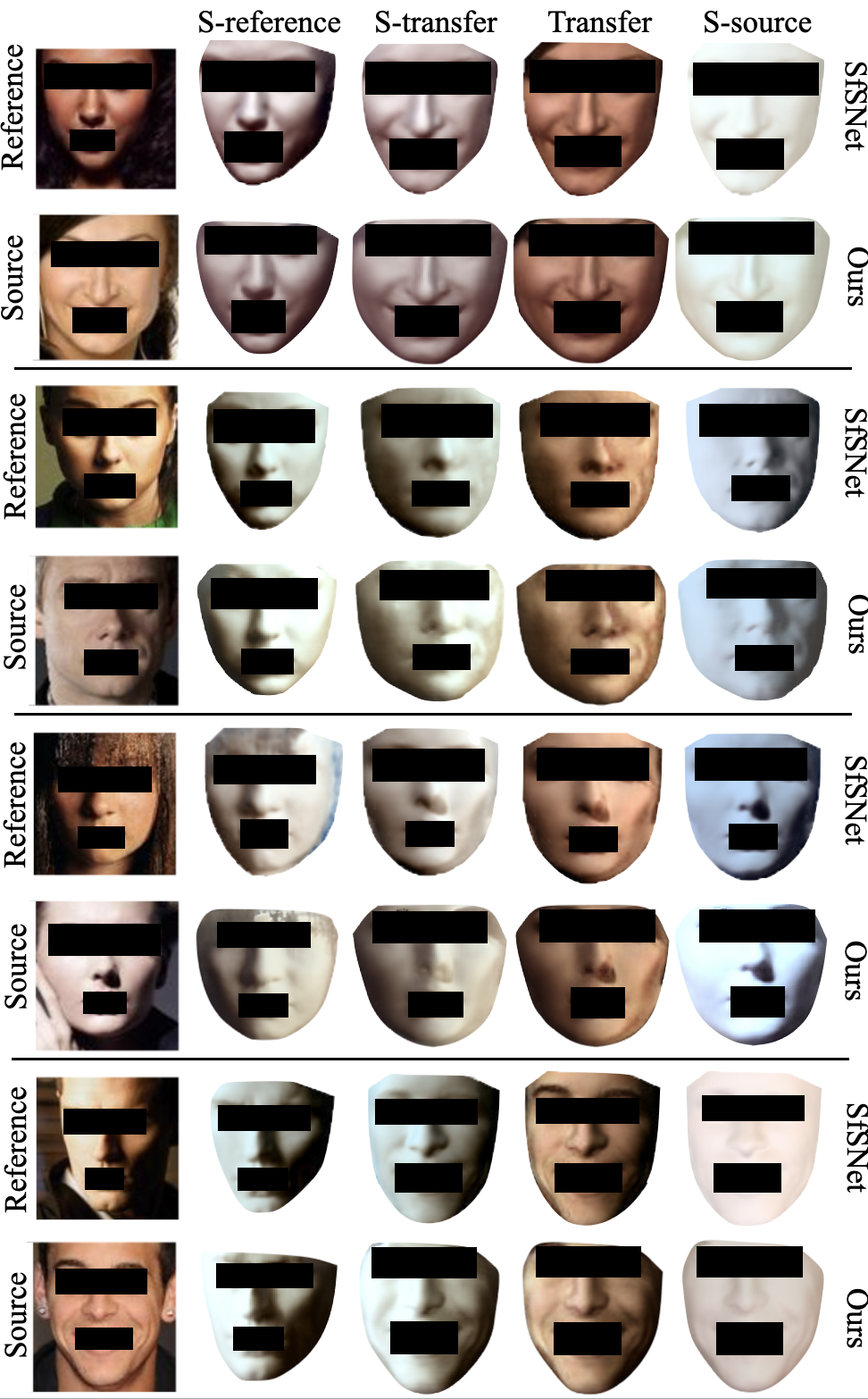}
\vspace{-15pt}
\caption{Results of light transfer and comparison with~\cite{sfsnetSengupta18}.}
\label{fig:light_transfer}
\vspace{-10pt}
\end{figure}

\begin{figure}
\centering
\includegraphics[width=1\linewidth]{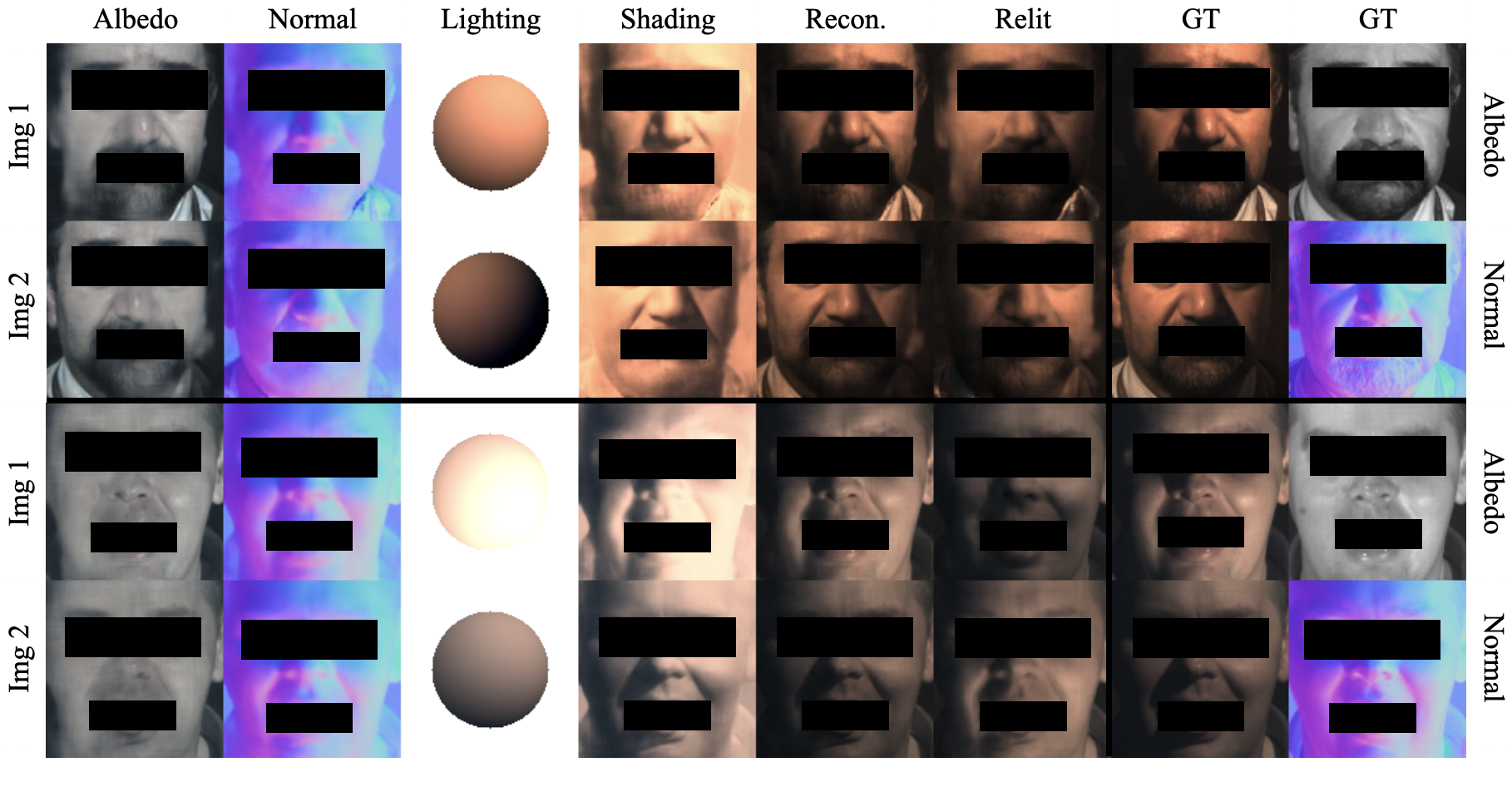}
\vspace{-20pt}
\caption{Visual decomposition and relighting results on Photoface dataset~\cite{photoface}.}
\vspace{-10pt}
\label{fig:photoface}
\end{figure}

\begin{figure}
\centering
\includegraphics[width=1\linewidth]{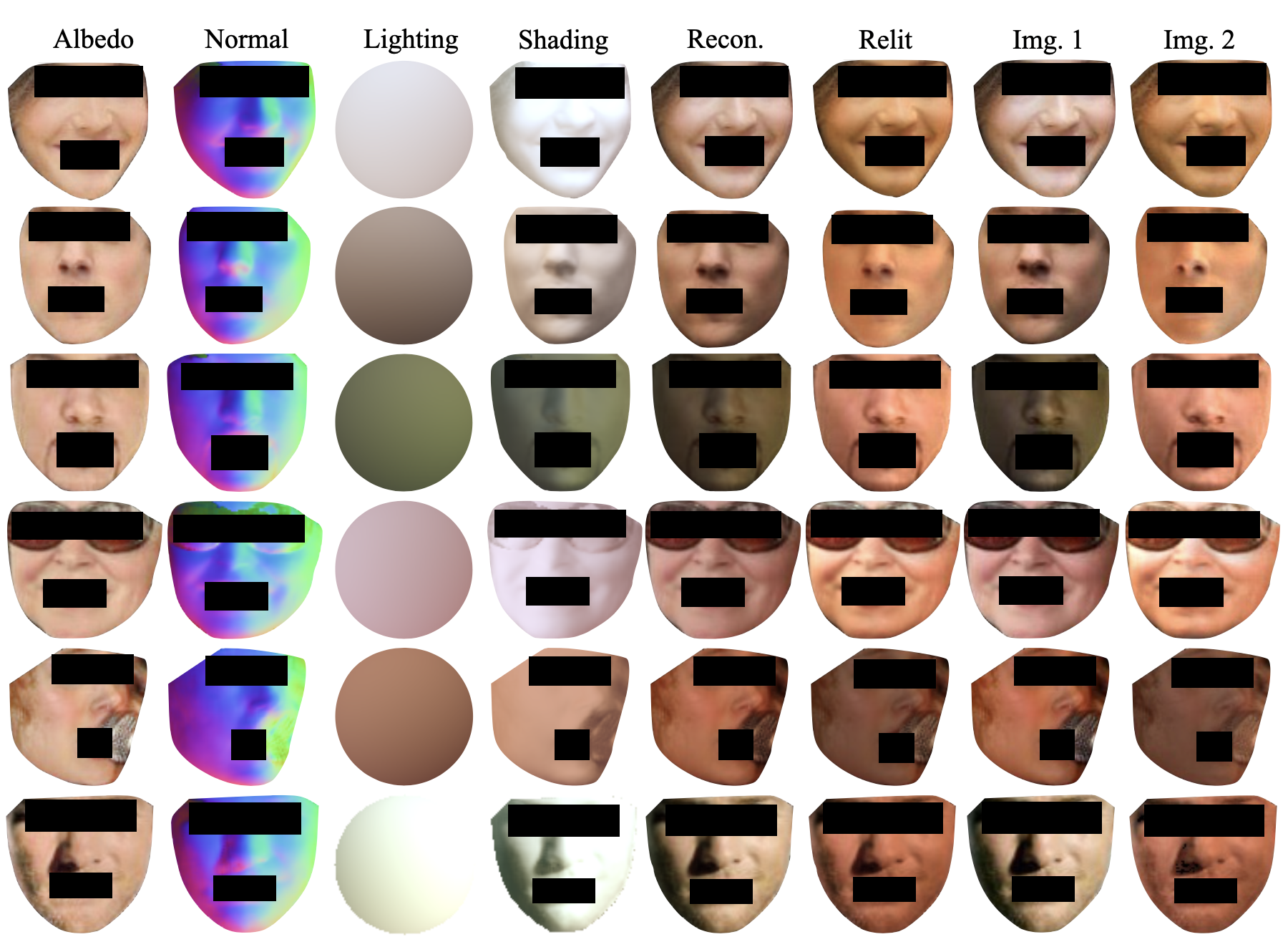}
\vspace{-10pt}
\caption{Quality of decomposition and relighting on mulit-lit CelebA dataset.}
\label{fig:decomp2}
\vspace{-10pt}
\end{figure}

\vspace{-3pt}
\subsection{Generating Mulit-Lit CelebA dataset}
\vspace{-3pt}
To generate the multi-lit CelebA dataset, we use the images from CelebA~\cite{liu2015faceattributes} and lighting vectors from synthetic SfSNet dataset~\cite{sfsnetSengupta18}. To this end, we train a skip-connection based encoder-decoder architecture (similar to~\cite{sfsnetSengupta18}) on the labeled synthetic dataset only. Then, we use this network to generate pseudo-labels for the CelebA dataset, while training another network with a mixture of SfSNet and CelebA datasets. Next, to relight CelebA images, we estimate their albedos and normals using the pretrained network, sample $5$ different lighting vectors from the SfSNet dataset and generate relit images by passing the estimated albedo, normal and sampled lightings through the rendering function in~\eqref{eq:lambertian}. During training with $\mathcal{L}_{\textrm{relit}}$, we choose random pairs of the relit images of the same subject. When training on a mix of synthetic and multi-lit CelebA dataset, to ensure that the original samples from the CelebA are used, we always choose the original image from CelebA as one of the images in the multi-lit image pair, while choosing the other image randomly from the relit versions. Samples of the multi-lit CelebA dataset are provided in Fig.~\ref{fig:dataset}-(a).
\vspace{-3pt}
\subsection{Training details}
\vspace{-3pt}
We examine two training paradigms: 1) train on a mix of synthetic and real datasets using the loss defined in~\eqref{eq:loss}, 2) train on multi-lit CelebA dataset with light supervision, $\mathcal{L}_{\textrm{rec}}$ and $\mathcal{L}_{\textrm{relit}}$. We adopt the synthetic dataset provided by~\cite{sfsnetSengupta18} which includes GT albedos, normals, SH lighting and segmentation maps alongside each image. A sample of this dataset is visualized in Fig.~\ref{fig:dataset}-(b). The SfSNet dataset consists of $\sim 190k$ samples, where we use $70\%$, $10\%$ and $20\%$ as train, validation and test splits, respectively.

For real unconstrained images, we use CelebA~\cite{liu2015faceattributes} and Photoface~\cite{photoface} datasets. To incorporate $\mathcal{L}_{\textrm{relit}}$ on CelebA during training, we use multi-lit CelebA dataset. The CelebA dataset originally contains $\sim 200k$ samples, but after generating its multi-lit version, $\sim 600k$ pairs of multi-lit images are available. We use the same splits provided in~\cite{liu2015faceattributes} for CelebA and multi-lit CelebA datasets. On the other hand, the Photoface dataset already contains images of the same subject and pose under different lightings. It also provides annotations in the forms of albedos and normals computed from shape from shading methods~\cite{sfs}. For each subject and pose, $4$ different lightings exist. Thus, during training we use ${4 \choose 2} = 6$ pairs of multi-lit images. To avoid having samples of the same subject in different splits, we separate the splits on a subject basis with $100$ identities in the test set (to further adhere to the existing baselines evaluated on this dataset). To preserve the personal identity of the subjects, we present results with redacted faces. We cover the eyes and mouth of the samples from real datasets with black boxes.

To further adapt CelebA images to the SfSNet dataset which only contain the face region, we extract the face region for each image. For this purpose, we create a face mask from the convex hull of the facial keypoints extracted by~\cite{dlib09} (outlined by the jaw line and the eyebrows). All images are then resized to $128 \times 128$. In the rest of this section, unless otherwise stated our models are trained on SfSNet and multi-lit CelebA dataset using the loss in~\eqref{eq:loss}.

In our experiments, we used two different encoder-decoder based architectures. In the first architecture (\textit{Arch 1}), we have skip connections between the encoder and both $A$ and $N$ decoders. We use Arch 1 for training models on SfSNet dataset only. In \textit{Arch 2}, we no longer have skip-connections. This architecture is used for training models on a mixture of SfSNet and real datasets. We use a similar network architecture as~\cite{sfsnetSengupta18} for Arch 1 and 2, following their ablation study and analysis on the effectiveness of each architecture in different settings. The light estimator module, consists of a $1 \! \times \! 1$ convolution layer followed by batch-norm, ReLU and adaptive average pooling that maps the features to a $256$-dimensional vector. This vector is then fed to a fully connected layer to output a $27$-dimensional lighting vector. In all experiments, we use the Adam optimizer with $5 \! \times \! 10^{-4}$ weight decay and learning rates between $0.0002$ and $0.001$ depending on the experiment. More details are provided in the supplementary material.

\begin{table}
\centering
\resizebox{\columnwidth}{!}{%
\begin{tabular}{ |c|c|c|c|c| }
\hline
Method & $\textrm{Mean} \pm \textrm{std}$ & $< 20^{\circ}$ & $< 25^{\circ}$ & $<30^{\circ}$ \\
\hline
3DMM & $26.3\pm10.2$ & $4.3 \%$ & $56.1\%$ & $89.4\%$ \\
Pix2Vertex~\cite{Sela_2017_ICCV} & $33.9\pm5.6$ & $24.8\%$ & $36.1\%$ &  $47.6\%$ \\
SfSNet(no ft)~\cite{sfsnetSengupta18} & $25.5\pm9.3$ & $43.6\%$ & $57.5\%$ &  $68.7\%$ \\
\hline
Marr Rev.~\cite{Bansal16} & $28.3\pm10.1$ & $31.8\%$ & $36.5\%$ &  $44.4\%$ \\
UberNet~\cite{Kokkinos_2017_CVPR} & $29.1\pm11.5$ & $30.8\%$ & $36.5\%$ &  $55.2\%$ \\
NiW~\cite{Trigeorgis_2017_cvpr} & $22.0\pm6.3$ & $36.6\%$ & $59.8\%$ &  $79.6\%$ \\
SfSNet-ft~\cite{sfsnetSengupta18} & $12.8\pm5.4$ & $83.7\%$ & $90.8\%$ &  $94.5\%$ \\
\hline
Ours-scratch & $\underline{12.22\pm9.2}$ & $\underline{85.3}\%$ & $\underline{91.9}\%$ &  $\underline{95.4}\%$ \\
Ours-ft & $\textbf{11.8}\pm\textbf{8.9}$ & $\textbf{86.3}\%$ & $\textbf{92.6}\%$ &  $\textbf{95.9}\%$ \\
\hline
\end{tabular}
}
\vspace{-10pt}
\caption{Comparison of angular error of the estimated normals on Photoface dataset. The best and second best results are highlighted in bold and underline.}
\label{tab:normal_results}
\vspace{-20pt}
\end{table}
\vspace{-3pt}
\subsection{Results}
\vspace{-3pt}
\noindent\textbf{Light transfer}: In Fig.~\ref{fig:relight_baselines}, we compare the light transfer quality of our model in contrast to several benchmarks on various datasets including MultiPIE~\cite{multipie}, DPR~\cite{DPR} and CelebA-HQ~\cite{CelebAMask-HQ}. Note that, we did not use samples from MultiPIE, DPR and CelebA-HQ datasets and rather took screenshots of the images provided in~\cite{DPR} and evaluated the performance of our models on these screenshots. We provide a brief description of the baselines as: DPR~\cite{DPR} trains an encoder-decoder architecture on multi-lit image pairs with lighting annotations, Shih et al~\cite{Shih2014} and Li et al.~\cite{Li_2018_ECCV} relight a content image given a style image, Shu et al~\cite{Shu2017} formulates the light transfer as a mass transport problem. Note that while these baselines have been trained on $512 \times 512$ images, our model and SfSNet~\cite{sfsnetSengupta18} are trained on $128 \times 128$ images. To have a fair comparison on our model and SfSNet, we resize the input images to $128 \times 128$ and operate on facial region, while reporting results for the other baselines on $512 \times 512$ images.

Note that, in~Fig.~\ref{fig:relight_baselines}-(a) and (c), the DPR baseline has the advantage of knowing the target lighting and relighting the source image based on this GT illumination. However, for our method and SfSNet we estimate the lighting from the reference and transfer it to the source image.
As our approach and SfSNet use colored light unlike DPR~\cite{DPR} (trained on monochromatic light), the color tone of the reference is transferred as the color of the light, thus the tone of the relit image matches the source. This is due to the nature of the synthetic dataset we have trained on, which contains colored light and low variety of skin tones. One can use other synthetic datasets with variations in the skin tones and monochromatic lighting. Note that our approach succeeds in accurately estimating the lighting from the reference image and relighting the source image even though we are training on lower resolution images without knowing the target lighting vector. Also, compared to the baselines, the transferred light better matches the reference and the relit image looks more natural with no visible artifacts (compared to the results from~\cite{Shih2014, Li_2018_ECCV, Shu2017}). These baselines suffer from accurately understanding the lighting from the reference image and then transferring it to the source image. Also in Fig.~\ref{fig:relight_baselines}-(a), compared to DPR, the facial shades on the relit images of our model better match the reference. This suggests that our model does a better job in estimating the lighting as it is trained on more natural looking images with a wide variation of lightings compared to the dataset used in DPR~\cite{DPR}. 

In Fig.~\ref{fig:light_transfer}, we closely examine the quality of the light transfer on CelebA dataset~\cite{liu2015faceattributes}. Note that compared to SfSNet, our model demonstrates more medium-frequency details (e.g. wrinkles) in the shading component (denoted by S-* in the figure). This in turn leads to better quality of the relit images by our model compared to SfSNet. We presume this is because of the use of pseudo labels during training by SfSNet. While these labels are helpful in stabilizing the training on real dataset, they might be erroneous and thus limit the quality of the learnt components. Furthermore, our method is implicitly trained on the light transfer task through the cross-relighting loss term, thus improving light transfer quality compared to SfSNet. In addition, Fig.~\ref{fig:light_transfer} exemplifies successful and challenging cases for the light transfer task. For instance, in the presence of cast shadows both our model and SfSNet have a hard time in accurate decomposition and thus transferring the light.

\noindent\textbf{Results on Photoface dataset}: We train our model with Arch 2 using $\mathcal{L}_{\textrm{rec}}$ and $\mathcal{L}_{\textrm{relit}}$ and  supervision on the normals and albedos (as they are provided in the Photoface dataset). We train on Photoface~\cite{photoface} with two different initializations of the network. In \textit{finetune}, as the name suggests, we finetune a network that is trained on the SfSNet and multi-lit CelebA dataset. In \textit{train from scratch}, we train a network from scratch using the Photoface dataset only. Visual results are provided in Fig.~\ref{fig:photoface}. Here, Img 1 and Img 2 denote images of the same subject with a fixed pose under different illuminations. The \textit{Relit} results are obtained by relighting Img 1 (Img 2) with the lighting from Img 2 (Img1). Thus, we expect the relit images to be close to Img 2 and Img1 respectively, as also observed in the figure. For the sake of brevity, here we only include visual results from \textit{finetune} training strategy and defer the results from \textit{train from scratch} to the supplementary material. Note that, both training strategies lead to reasonable results in terms of decomposition and relighting, hence suggesting the effectiveness of the cross-relighting loss term in training networks from scratch. Furthermore, it is observed that the decomposition and the relighting task on parts of the face which are under strong shadows or dimly lit is more challenging as they contain less information.

In addition, we assess the quality of the estimated normals by our method against several baselines trained on Photoface in Table~\ref{tab:normal_results}. The metric used is the angular error (in degrees) between the GT and estimated normals at each pixel. We report on mean (per pixel error) and the percentage of pixels with normal angular errors less than $20^{\circ}$, $25^{\circ}$ and $30^{\circ}$. Note that, our method outperforms all previous methods in terms of the quality of the estimated normals. This gain verifies the importance of using $\mathcal{L}_{\textrm{relit}}$ and how it helps in a more accurate decomposition especially for low-light images. 

\begin{table}
\setlength{\tabcolsep}{2pt}
\centering
\begin{tabular}{ |c|c|c|c|c|c| }
\hline
\footnotesize{Metric/Model} & \footnotesize{DPR~\cite{DPR}} & \footnotesize{Pix2Pix~\cite{pix2pix2017}} & \footnotesize{Syn. only~\cite{sfsnetSengupta18}} & \footnotesize{SfSNet~\cite{sfsnetSengupta18}} & \footnotesize{Ours} \\
\hline
\footnotesize{$L_1$ recon} & 26.7 & 30.9 & 42.2 & 26.3 & \textbf{17.8}  \\
\footnotesize{$L_1$ relit} & 34.9 & 48.7 & 48.7 &  39.2 & \textbf{30.8}  \\
\hline
\footnotesize{$L_2$ recon} & 20.8 & 25.5 & 32.4 & 19.9 & \textbf{12.8} \\
\footnotesize{$L_2$ relit} & 27.3 & 37.0 & 37.4 & 30.4 & \textbf{23.5} \\
\hline
\footnotesize{SSIM recon} & 0.94 & 0.95 & 0.92 & 0.95 & \textbf{0.97} \\
\footnotesize{SSIM relit} & 0.9 & 0.89 & 0.89 & 0.92 & \textbf{0.93} \\
\hline
\end{tabular}
\vspace{-10pt}
\caption{Quantitative comparison with DPR and Pix2Pix on reconstruction and relighting tasks.}
\label{tab:dpr_pix_results}
\vspace{-20pt}
\end{table}

\begin{table*}
\centering
\begin{tabular}{ |c|c|c|c|c|c|c|c| }
\hline
Model / Metric & Decomp. & $L_1$ recon & $L_1$ relit & $L_2$ recon & $L_2$ relit & SSIM recon & SSIM relit \\
\hline
\footnotesize{Syn. + Real + No Pseudo sp} & \xmark & 12.2 & 128.6 & 14.5 & 92.8 & 0.97 & 0.83 \\
\footnotesize{Syn. only} & \cmark & 42.2 & 48.7 & 32.4 & 37.4 & 0.92 & 0.89\\
\footnotesize{Syn. + Real + Pseudo sp} & \cmark & 26.3 & 39.2 & 19.9 & 30.4 & 0.94 & 0.92 \\
\footnotesize{Syn. + Real + CR} & \cmark & 17.3 & 32.0 & 12.5 & 24.4 & 0.97 & 0.93 \\
\footnotesize{Syn. + Real + CR + LAB} & \cmark & 17.8 & 30.9 & 12.8 & 23.5 & 0.97 & 0.93 \\
\hline
\footnotesize{Real + light} & \xmark & 19.7 & 46.7 & 15.0 & 37.7 & 0.96 & 0.88 \\
\footnotesize{Real + CR + light} & \cmark & 17.8 & 25.8 & 13.8 & 20.4 & 0.96 & 0.93 \\
\hline
\end{tabular}
\caption{Quantitative evaluation on multi-lit CelebA dataset. The second row contains the results of a model trained only on the synthetic data, while the rest are trained on a mix of SfSNet and multi-lit CelebA datasets. (Pseudo sp: pseudo supervision, CR: cross-relighting).}
\label{tab:results}
\end{table*}

\noindent\textbf{Decomposition performance}: In Fig.~\ref{fig:decomp2}, we present the estimated intrinsic components by our model. We also relight Img. 1 with the estimated lighting from Img. 2, hence we expect the relit image to be close to Img. 2. Our results verify the accuracy in the estimated components and highlight challenging scenarios for the decomposition task (last three examples). While the model generalizes well on various poses, it is having difficulty in the intrinsic decomposition task on parts of the images which are partially occluded or contain cast shadows. This is due to the scarcity of such examples with GT labels in the training set.

\begin{figure}
\centering
\includegraphics[width=1\linewidth]{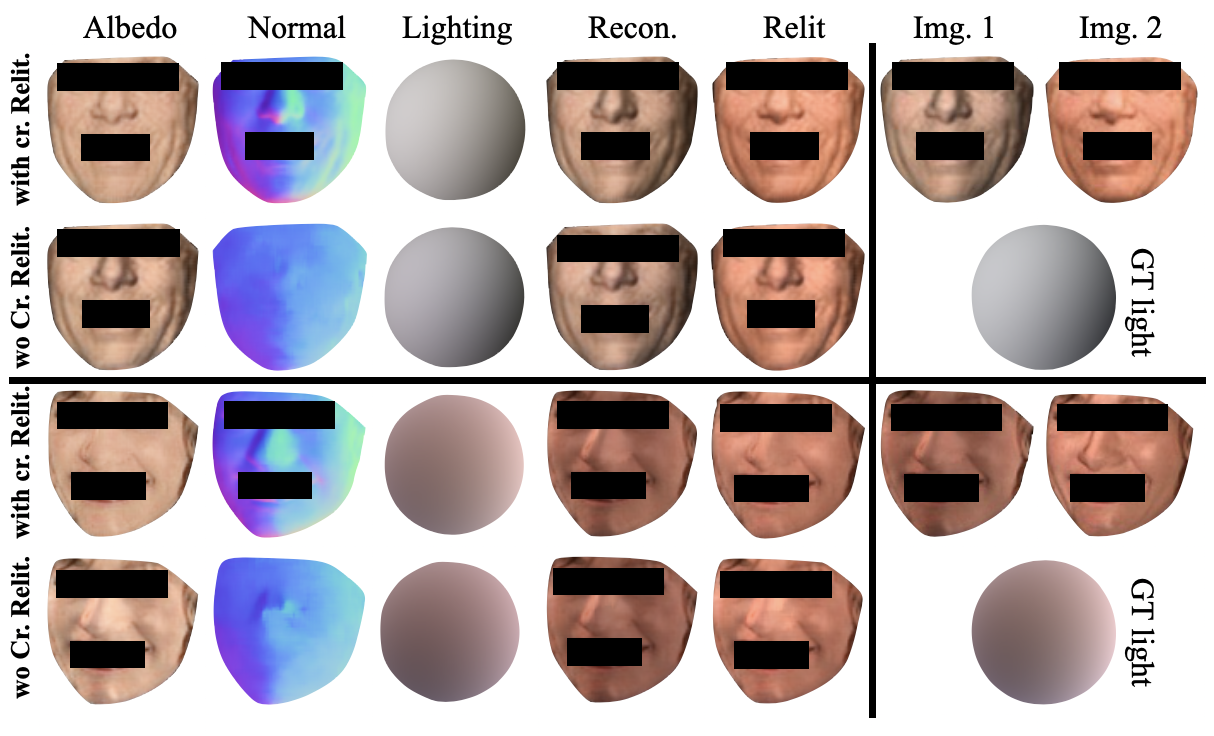}
\vspace{-10pt}
\caption{Quality of decomposition and relighting of models trained only on multi-lit CelebA dataset with light and reconstruction supervision, with and without $\mathcal{L}_{\textrm{relit}}$.}
\label{fig:light_exp}
\vspace{-10pt}
\end{figure}

\noindent\textbf{Importance of cross-relighting loss}: To highlight the importance of $\mathcal{L}_{\textrm{relit}}$, we train two models only on multi-lit CelebA dataset. In both models we train with light supervision and $\mathcal{L}_{\textrm{rec}}$. We also use $\mathcal{L}_{\textrm{relit}}$ while training the second model. Note that no form of supervision on $A$ and $N$ are provided in training these two models. In Fig. ~\ref{fig:light_exp}, we provide visual results on the quality of the decomposition and relighting. While both models are successful in the lighting estimation, the model trained without $\mathcal{L}_{\textrm{relit}}$ clearly fails in decomposing the albedo and normals. Even though, it still provides reasonable looking relit images in terms of the color tone, it evidently misses the correct shading which is a result of the interaction between lighting and normals. This stems from the ambiguity of the decomposition task, especially if no reference albedos and normals are provided (even from synthetic data). However, using $\mathcal{L}_{\textrm{relit}}$ compensates this lack of supervision and results in accurate decomposition and relighting. Our quantitative results in Table~\ref{tab:results} (last two rows) also verify this.

\noindent\textbf{Ablation study}: In Table~\ref{tab:results}, we present a quantitative ablation study on the test set of the multi-lit CelebA dataset. Our metrics are average per pixel $L_1$ and $L_2$ errors for the reconstructed and relit images alongside structural similarity index measure (SSIM). While there is no GT labels for the albedos and normals for CelebA multi-lit dataset, we evaluate the decomposition performance visually. For training on synthetic data only i.e. \textit{Syn. Only}, we train a network with Arch 1 using the SfSNet dataset. For the rest of the methods, we train a network with Arch 2. Pseudo sp (pseudo supervision) refers to the use of pseudo-labels obtained from the pre-trained \textit{Syn. Only} model during training. The results demonstrate that without pseudo-supervision or the use of $\mathcal{L}_{\textrm{relit}}$, learning intrinsic decomposition from a mix of synthetic and real datasets fail. Furthermore, the results verify the improvement provided by imposing $\mathcal{L}_{\textrm{rec}}$ and $\mathcal{L}_{\textrm{relit}}$ in LAB color space. 

In the last two rows, we quantify the performance of the model trained only on multi-lit CelebA dataset with light and reconstruction supervision and with/without $\mathcal{L}_{\textrm{relit}}$. The results again verify the importance of the cross-relighting loss in limited supervision settings and how a model trained with light, $\mathcal{L}_{\textrm{rec}}$ and $\mathcal{L}_{\textrm{relit}}$ on multi-lit CelebA dataset has close (or even better on some metrics) performance to the models trained on a hybrid dataset.

\noindent\textbf{Comparison to DPR and Pix2Pix}: We compare our model against two benchmarks specialized for the subject relighting task and retrained on multi-lit CelebA. For DPR~\cite{DPR}, we train a skip-connection based encoder-decoder architecture which relights an input image given a target illumination. In Pix2Pix~\cite{pix2pix2017}, we treat the lighting as a style vector and train a generator-discriminator pair adversarially, conditioned on the lighting vector. Note that for these two baselines, the target lighting is required and assumed to be known during training unlike our model. We report the quantitative results in Table~\ref{tab:dpr_pix_results} where we have retrained SfSNet on multi-lit CelebA. Note that our model outperforms all baselines on both reconstruction and relighting tasks. Compared to DPR and Pix2Pix which only focus on the relighting task and use the lighting vector alongside the image/features to relight an image, our model additionally performs intrinsic decomposition which is helpful in improving the relighting performance.

\section{Conclusion}
\label{sec:conclusion}
\vspace{-5pt}
In this paper, we studied the inverse rendering problem, i.e. decomposing a single view image into its intrinsic components such as albedo, normals and illumination. We proposed a new self-supervised training paradigm, leveraging consistencies between images of the same scene under different illuminations, i.e. multi-lit images. We also generated a multi-lit dataset from the CelebA dataset, named multi-lit CelebA. We examined our training approach under two different settings: 1) hybrid training on synthetic and multi-lit CelebA dataset, 2) limited supervision training on multi-lit CelebA. We showed in both settings our model is successful in learning intrinsic decomposition and relighting tasks despite the challenging ambiguities of the problem. We presented qualitative and quantitative comparisons with various state-of-the-art baselines to demonstrate the strength of our approach.

{\small
\bibliographystyle{ieee_fullname}
\bibliography{refs.bib}

\begin{thebibliography}{10}\itemsep=-1pt

\bibitem{Bansal16}
A. Bansal, B. Russell, and A. Gupta.
\newblock Marr {R}evisited: 2{D}-3{D} model alignment via surface normal
  prediction.
\newblock In {\em CVPR}, 2016.

\bibitem{Malik2015}
J.~T. {Barron} and J. {Malik}.
\newblock Shape, illumination, and reflectance from shading.
\newblock {\em IEEE Transactions on Pattern Analysis and Machine Intelligence},
  37(8):1670--1687, 2015.

\bibitem{barrow1978recovering}
H. Barrow, J. Tenenbaum, A. Hanson, and E. Riseman.
\newblock Recovering intrinsic scene characteristics.
\newblock {\em Comput. Vis. Syst}, 2(3-26):2, 1978.

\bibitem{Baslamisli_2018_ECCV}
A.~S. Baslamisli, T.~T. Groenestege, P. Das, H. Le, S. Karaoglu, and T. Gevers.
\newblock Joint learning of intrinsic images and semantic segmentation.
\newblock In {\em Proceedings of the European Conference on Computer Vision
  (ECCV)}, September 2018.

\bibitem{3DMM}
V. Blanz and T. Vetter.
\newblock A morphable model for the synthesis of 3d faces.
\newblock In {\em Proceedings of the 26th Annual Conference on Computer
  Graphics and Interactive Techniques}, SIGGRAPH '99, page 187–194, USA,
  1999. ACM Press/Addison-Wesley Publishing Co.

\bibitem{chen2019photo}
A. Chen, Z. Chen, G. Zhang, K. Mitchell, and J. Yu.
\newblock Photo-realistic facial details synthesis from single image.
\newblock In {\em Proceedings of the IEEE International Conference on Computer
  Vision}, pages 9429--9439, 2019.

\bibitem{chen2020neural}
Z. Chen, A. Chen, G. Zhang, C. Wang, Y. Ji, K.~N. Kutulakos, and J. Yu.
\newblock A neural rendering framework for free-viewpoint relighting.
\newblock In {\em Proceedings of the IEEE/CVF Conference on Computer Vision and
  Pattern Recognition}, pages 5599--5610, 2020.

\bibitem{choy2016}
C.~B. Choy, D. Xu, J. Gwak, K. Chen, and S. Savarese.
\newblock 3d-r2n2: A unified approach for single and multi-view 3d object
  reconstruction.
\newblock In Bastian Leibe, Jiri Matas, Nicu Sebe, and Max Welling, editors,
  {\em Computer Vision -- ECCV 2016}, pages 628--644, Cham, 2016. Springer
  International Publishing.

\bibitem{Debevec2000}
P. Debevec, T. Hawkins, C. Tchou, H. Duiker, W. Sarokin, and M. Sagar.
\newblock Acquiring the reflectance field of a human face.
\newblock In {\em Proceedings of the 27th Annual Conference on Computer
  Graphics and Interactive Techniques}, SIGGRAPH '00, page 145–156, USA,
  2000. ACM Press/Addison-Wesley Publishing Co.

\bibitem{sfs}
R.~T. {Frankot} and R. {Chellappa}.
\newblock A method for enforcing integrability in shape from shading
  algorithms.
\newblock {\em IEEE Transactions on Pattern Analysis and Machine Intelligence},
  10(4):439--451, 1988.

\bibitem{Debevec2015}
G. {Fyffe} and P. {Debevec}.
\newblock Single-shot reflectance measurement from polarized color gradient
  illumination.
\newblock In {\em 2015 IEEE International Conference on Computational
  Photography (ICCP)}, pages 1--10, 2015.

\bibitem{Fyffe2009}
G. {Fyffe}, C.~A. {Wilson}, and P. {Debevec}.
\newblock Cosine lobe based relighting from gradient illumination photographs.
\newblock In {\em 2009 Conference for Visual Media Production}, pages 100--108,
  2009.

\bibitem{multipie}
R. Gross, I. Matthews, J. Cohn, T. Kanade, and S. Baker.
\newblock Multi-pie.
\newblock 2013.

\bibitem{bn}
S. Ioffe and C. Szegedy.
\newblock Batch normalization: Accelerating deep network training by reducing
  internal covariate shift.
\newblock In {\em Proceedings of the 32nd International Conference on
  International Conference on Machine Learning - Volume 37}, ICML'15, page
  448–456. JMLR.org, 2015.

\bibitem{pix2pix2017}
P. Isola, J. Zhu, T. Zhou, and A.~A Efros.
\newblock Image-to-image translation with conditional adversarial networks.
\newblock {\em CVPR}, 2017.

\bibitem{janner2017self}
M. Janner, J. Wu, T.~D Kulkarni, I. Yildirim, and J. Tenenbaum.
\newblock Self-supervised intrinsic image decomposition.
\newblock In {\em Advances in Neural Information Processing Systems}, pages
  5936--5946, 2017.

\bibitem{CelebAMask-HQ}
T. Karras, T. Aila, S. Laine, and J. Lehtinen.
\newblock Progressive growing of {GAN}s for improved quality, stability, and
  variation.
\newblock In {\em International Conference on Learning Representations}, 2018.

\bibitem{Kim_2016_CVPR}
J. Kim, J. Kwon~Lee, and K. Mu~Lee.
\newblock Accurate image super-resolution using very deep convolutional
  networks.
\newblock In {\em Proceedings of the IEEE Conference on Computer Vision and
  Pattern Recognition (CVPR)}, June 2016.

\bibitem{dlib09}
D.~E. King.
\newblock Dlib-ml: A machine learning toolkit.
\newblock {\em Journal of Machine Learning Research}, 10:1755--1758, 2009.

\bibitem{Kokkinos_2017_CVPR}
I. Kokkinos.
\newblock Ubernet: Training a universal convolutional neural network for low-,
  mid-, and high-level vision using diverse datasets and limited memory.
\newblock In {\em Proceedings of the IEEE Conference on Computer Vision and
  Pattern Recognition (CVPR)}, July 2017.

\bibitem{kulkarni2015deep}
T.~D Kulkarni, W.~F Whitney, P. Kohli, and J. Tenenbaum.
\newblock Deep convolutional inverse graphics network.
\newblock In {\em Advances in neural information processing systems}, pages
  2539--2547, 2015.

\bibitem{land1971lightness}
E.~H Land and J.~J McCann.
\newblock Lightness and retinex theory.
\newblock {\em Josa}, 61(1):1--11, 1971.

\bibitem{Lettry2018}
L. {Lettry}, K. {Vanhoey}, and L. {van Gool}.
\newblock Darn: A deep adversarial residual network for intrinsic image
  decomposition.
\newblock In {\em 2018 IEEE Winter Conference on Applications of Computer
  Vision (WACV)}, pages 1359--1367, 2018.

\bibitem{Li_2018_ECCV}
Y. Li, M. Liu, X. Li, M. Yang, and J. Kautz.
\newblock A closed-form solution to photorealistic image stylization.
\newblock In {\em Proceedings of the European Conference on Computer Vision
  (ECCV)}, September 2018.

\bibitem{BigTimeLi18}
Z. Li and N. Snavely.
\newblock Learning intrinsic image decomposition from watching the world.
\newblock In {\em Computer Vision and Pattern Recognition (CVPR)}, 2018.

\bibitem{li2018learning}
Z. Li, Z. Xu, R. Ramamoorthi, K. Sunkavalli, and M. Chandraker.
\newblock Learning to reconstruct shape and spatially-varying reflectance from
  a single image.
\newblock In {\em SIGGRAPH Asia 2018 Technical Papers}, page 269. ACM, 2018.

\bibitem{liu2015faceattributes}
Z. Liu, P. Luo, X. Wang, and X. Tang.
\newblock Deep learning face attributes in the wild.
\newblock In {\em Proceedings of International Conference on Computer Vision
  (ICCV)}, December 2015.

\bibitem{ma2018single}
Wei-Chiu Ma, Hang Chu, Bolei Zhou, Raquel Urtasun, and Antonio Torralba.
\newblock Single image intrinsic decomposition without a single intrinsic
  image.
\newblock In {\em Proceedings of the European Conference on Computer Vision
  (ECCV)}, pages 201--217, 2018.

\bibitem{Meka2019}
A. Meka, C. Haene, R. Pandey, M. Zollhoefer, S. Fanello, G. Fyffe, A. Kowdle,
  X. Yu, J. Busch, J. Dourgarian, P. Denny, S. Bouaziz, P. Lincoln, M. Whalen,
  G. Harvey, J. Taylor, S. Izadi, A. Tagliasacchi, P. Debevec, C. Theobalt, J.
  Valentin, and C. Rhemann.
\newblock Deep reflectance fields - high-quality facial reflectance field
  inference from color gradient illumination.
\newblock volume~38, July 2019.

\bibitem{cgan}
Mehdi {Mirza} and Simon {Osindero}.
\newblock {Conditional Generative Adversarial Nets}.
\newblock {\em arXiv e-prints}, page arXiv:1411.1784, Nov. 2014.

\bibitem{Narihira_2015}
T. Narihira, M. Maire, and S.~X. Yu.
\newblock Direct intrinsics: Learning albedo-shading decomposition by
  convolutional regression.
\newblock In {\em Proceedings of the IEEE International Conference on Computer
  Vision (ICCV)}, December 2015.

\bibitem{Nestmeyer_2020_CVPR}
T. Nestmeyer, J. Lalonde, I. Matthews, and A. Lehrmann.
\newblock Learning physics-guided face relighting under directional light.
\newblock In {\em Proceedings of the IEEE/CVF Conference on Computer Vision and
  Pattern Recognition (CVPR)}, June 2020.

\bibitem{Ramamoorthi2001}
R. Ramamoorthi and P. Hanrahan.
\newblock An efficient representation for irradiance environment maps.
\newblock In {\em Proceedings of the 28th Annual Conference on Computer
  Graphics and Interactive Techniques}, SIGGRAPH ’01, page 497–500, New
  York, NY, USA, 2001. Association for Computing Machinery.

\bibitem{Zhang1999}
{Ruo Zhang}, {Ping-Sing Tsai}, J.~E. {Cryer}, and M. {Shah}.
\newblock Shape-from-shading: a survey.
\newblock {\em IEEE Transactions on Pattern Analysis and Machine Intelligence},
  21(8):690--706, 1999.

\bibitem{Sela_2017_ICCV}
M. Sela, E. Richardson, and R. Kimmel.
\newblock Unrestricted facial geometry reconstruction using image-to-image
  translation.
\newblock In {\em Proceedings of the IEEE International Conference on Computer
  Vision (ICCV)}, Oct 2017.

\bibitem{neuralSengupta19}
S. Sengupta, J. Gu, K. Kim, G. Liu, D.~W. Jacobs, and J. Kautz.
\newblock Neural inverse rendering of an indoor scene from a single image.
\newblock In {\em International Conference on Computer Vision (ICCV)}, 2019.

\bibitem{sfsnetSengupta18}
S. Sengupta, A. Kanazawa, C.~D. Castillo, and D.~W. Jacobs.
\newblock Sfsnet: Learning shape, refectance and illuminance of faces in the
  wild.
\newblock In {\em Computer Vision and Pattern Regognition (CVPR)}, 2018.

\bibitem{shi2017learning}
J. Shi, Y. Dong, H. Su, and S.~X Yu.
\newblock Learning non-lambertian object intrinsics across shapenet categories.
\newblock In {\em Proceedings of the IEEE Conference on Computer Vision and
  Pattern Recognition}, pages 1685--1694, 2017.

\bibitem{Shih2014}
Y. Shih, S. Paris, C. Barnes, W.~T. Freeman, and F. Durand.
\newblock Style transfer for headshot portraits.
\newblock {\em ACM Trans. Graph.}, 33(4), July 2014.

\bibitem{Shu2017}
Z. Shu, S. Hadap, E. Shechtman, K. Sunkavalli, S. Paris, and D. Samaras.
\newblock Portrait lighting transfer using a mass transport approach.
\newblock {\em ACM Trans. Graph.}, 37(1), Oct. 2017.

\bibitem{NeuralFace2017}
Z. Shu, E. Yumer, S. Hadap, K. Sunkavalli, E. Shechtman, and D. Samaras.
\newblock Neural face editing with intrinsic image disentangling.
\newblock In {\em Computer Vision and Pattern Recognition (CVPR)}. IEEE, 2017.

\bibitem{song2020recovering}
G. Song, J. Zheng, J. Cai, and T. Cham.
\newblock Recovering facial reflectance and geometry from multi-view images.
\newblock {\em Image and Vision Computing}, page 103897, 2020.

\bibitem{Sun2019}
T. Sun, J.~T. Barron, Y. Tsai, Z. Xu, X. Yu, G. Fyffe, C. Rhemann, J. Busch, P.
  Debevec, and R. Ramamoorthi.
\newblock Single image portrait relighting.
\newblock {\em ACM Trans. Graph.}, 38(4), July 2019.

\bibitem{tang2012deep}
Y. Tang, R. Salakhutdinov, and G. Hinton.
\newblock Deep lambertian networks.
\newblock 2012.

\bibitem{tappen2005}
M.~F. {Tappen}, W.~T. {Freeman}, and E.~H. {Adelson}.
\newblock Recovering intrinsic images from a single image.
\newblock {\em IEEE Transactions on Pattern Analysis and Machine Intelligence},
  27(9):1459--1472, 2005.

\bibitem{mofa}
A. Tewari, M. Zollhofer, H. Kim, P. Garrido, F. Bernard, P. Perez, and C.
  Theobalt.
\newblock Mofa: Model-based deep convolutional face autoencoder for
  unsupervised monocular reconstruction.
\newblock In {\em Proceedings of the IEEE International Conference on Computer
  Vision (ICCV) Workshops}, Oct 2017.

\bibitem{Trigeorgis_2017_cvpr}
G. {Trigeorgis}, P. {Snape}, I. {Kokkinos}, and S. {Zafeiriou}.
\newblock Face normals "in-the-wild" using fully convolutional networks.
\newblock In {\em 2017 IEEE Conference on Computer Vision and Pattern
  Recognition (CVPR)}, pages 340--349, 2017.

\bibitem{trigeorgis2017face}
G. Trigeorgis, P. Snape, I. Kokkinos, and S. Zafeiriou.
\newblock Face normals “in-the-wild” using fully convolutional networks.
\newblock In {\em Proceedings of the IEEE Conference on Computer Vision and
  Pattern Recognition}, pages 38--47, 2017.

\bibitem{Wu_2018_ECCV}
J. Wu, C. Zhang, X. Zhang, Z. Zhang, W.~T. Freeman, and J.~B. Tenenbaum.
\newblock Learning shape priors for single-view 3d completion and
  reconstruction.
\newblock In {\em Proceedings of the European Conference on Computer Vision
  (ECCV)}, September 2018.

\bibitem{relu}
B. {Xu}, N. {Wang}, T. {Chen}, and Mu {Li}.
\newblock {Empirical Evaluation of Rectified Activations in Convolutional
  Network}.
\newblock {\em arXiv e-prints}, page arXiv:1505.00853, May 2015.

\bibitem{xu2019deep}
Z. Xu, S. Bi, K. Sunkavalli, S. Hadap, H. Su, and R. Ramamoorthi.
\newblock Deep view synthesis from sparse photometric images.
\newblock {\em ACM Transactions on Graphics (TOG)}, 38(4):1--13, 2019.

\bibitem{xu2018}
Z. Xu, K. Sunkavalli, S. Hadap, and R. Ramamoorthi.
\newblock Deep image-based relighting from optimal sparse samples.
\newblock {\em ACM Trans. Graph.}, 37(4), July 2018.

\bibitem{Yeh_2017_CVPR}
R.~A. Yeh, C. Chen, T. Yian~Lim, A.~G. Schwing, M. Hasegawa-Johnson, and M.~N.
  Do.
\newblock Semantic image inpainting with deep generative models.
\newblock In {\em Proceedings of the IEEE Conference on Computer Vision and
  Pattern Recognition (CVPR)}, July 2017.

\bibitem{yu19inverserendernet}
Y. Yu and W.~AP Smith.
\newblock Inverserendernet: Learning single image inverse rendering.
\newblock In {\em Proceedings of the IEEE/CVF Conference on Computer Vision and
  Pattern Recognition (CVPR)}, 2019.

\bibitem{photoface}
S. Zafeiriou, M. Hansen, G. Atkinson, V. Argyriou, M. Petrou, M. Smith, and L.
  Smith.
\newblock The photoface database.
\newblock {\em IEEE Computer Society Conference on Computer Vision and Pattern
  Recognition Workshops}, pages 132--139, 2011.

\bibitem{zhang2020neural}
X. Zhang, S. Fanello, Y. Tsai, T. Sun, T. Xue, R. Pandey, S. Orts-Escolano, P.
  Davidson, C. Rhemann, P. Debevec, J.~T. Barron, R. Ramamoorthi, and W.~T.
  Freeman.
\newblock Neural light transport for relighting and view synthesis.
\newblock {\em arXiv preprint arXiv:2008.03806}, 2020.

\bibitem{DPR}
H. Zhou, S. Hadap, K. Sunkavalli, and D.~W. Jacobs.
\newblock Deep single portrait image relighting.
\newblock In {\em International Conference on Computer Vision (ICCV)}, 2019.

\bibitem{zhou2018label}
H. Zhou, J. Sun, Y. Yacoob, and D.~W Jacobs.
\newblock Label denoising adversarial network (ldan) for inverse lighting of
  faces.
\newblock In {\em Proceedings of the IEEE Conference on Computer Vision and
  Pattern Recognition}, pages 6238--6247, 2018.

\bibitem{Zhu2020}
X. {Zhu}, X. {Han}, W. {Zhang}, J. {Zhao}, and L. {Liu}.
\newblock Learning intrinsic decomposition of complex-textured fashion images.
\newblock In {\em 2020 IEEE International Conference on Multimedia and Expo
  (ICME)}, pages 1--6, 2020.

\end{thebibliography}
}

\section{Supplementary Material}
This section includes supplementary results and descriptions of the experiments discussed in the main draft.
\label{sec:appendix}

\subsection{Notations}
Let $\textrm{Conv}(c_i, c_o, k, s, p)$ denote a convolution layer with $c_i$ and $c_o$ input and output channels, $k$ kernel size, stride $s$ and padding $p$. Also, $\textrm{bn}(c_i)$ represents a batch normalization layer with $c_i$ input channels. We define a convolution layer followed by ReLU and batchnorm as a convolution block, denoted by $\textrm{ConvB}(c_i, c_o, k, s, p) = [\textrm{Conv}(c_i, c_o, k, s, p)- \textrm{relu}-\textrm{bn}(c_o)]$. We also define deconvolution block similarly as $\textrm{DeconvB}(c_i, c_o, k, s, p) = [\textrm{ConvTranspose}(c_i, c_o, k, s, p, p_o)- \textrm{relu}-\textrm{bn}(c_o)]$ with $p_o$ denoting the output padding.  Also, $FC(n_i, n_o)$ is a fully connected layer with $n_i$ and $n_o$ input and output sizes. 

\subsection{Our architecture}
We use two architectures in our experiments. In Arch 1 which is used for training on the synthetic dataset only, we have a U-net based encoder-decoder. The architecture of the encoder is $[\textrm{ConvB}(3, 32, 5, 2, 2)-\textrm{ConvB}(32, 64, 5, 2, 2)-\textrm{ConvB}(64, 128, 5, 2, 2)-\textrm{ConvB}(128, 256, 5, 2, 2)]$. The features output by the encoder are then passed through two separate cascade of convolution blocks ($[\textrm{ConvB}(256, 256, 3, 1, 1)-\textrm{ConvB}(256, 256, 3, 1, 1)-\textrm{ConvB}(256, 256, 3, 1, 1)]$), to represent the normal and albedo features. The normal and albedo features are then passed through two separate decoders with the same architectures as $[\textrm{DeconvB}(256, 128, 5, 2, 2, 1)-\textrm{DeconvB}(128, 64, 5, 2, 2, 1)-\textrm{DeconvB}(64, 32, 5, 2, 2, 1)-\textrm{DeconvB}(32, 16, 5, 2, 2, 1)-\textrm{Conv}(16, 3, 5, 2, 1)]$. Note that we have skip-connections from different layers of the encoder to the respective layers of the decoders. For the light decoder, we concatenate the normal and albedo features along with the image features from the encoder and pass it through a $\textrm{ConvB}(768, 256, 1, 1, 0)$, an average pooling and $FC(256, 27)$. 

In Arch 2, which is used for training on a combination of synthetic and real dataset, we no longer have skip connections. We use the same architecture as~\cite{sfsnetSengupta18} for Arch 2.

\subsection{Details on the DPR baseline}
Our architecture for the DPR baseline is a skip-connection based encoder-decoder similar to~\cite{DPR}. The encoder consists of $5$ convolution blocks as  $[\textrm{ConvB}(3, 16, 5, 1, 2)\!-\!\textrm{ConvB}(16, 16, 3, 2, 1)-\textrm{ConvB}(16, 32, 3, 2, 1)-\textrm{ConvB}(32, 64, 3, 2, 1)-\textrm{ConvB}(64, 155, 3, 2, 1)]$. The last $27$ channels of the extracted feature from the encoder is then passed through an adaptive average pooling layer followed by two fully connected layers to output a $27 \times 1$ lighting vector. This would be the estimated lighting vector for the input (source) image.

At the decoder side, the target lighting is first passed through two fully connected layers (followed by batch norm and ReLU) to generate a $27 \times 1$ feature vector for the target lighting. This feature vector is then repeated to have the same width and height as the 2D features generated by the encoder. This repeated feature is then concatenated with the first $128$ channels of the encoder feature to form a feature with $155$ channels. This feature is then passed through the decoder structured as: $[\textrm{DeconvB}(155, 64, 3, 2, 1, 1)-\textrm{DeconvB}(64, 32, 3, 2, 1, 1)-\textrm{DeconvB}(32, 16, 3, 2, 1, 1)-\textrm{DeconvB}(16, 16, 3, 2, 1, 1)-\textrm{Conv}(16, 3, 3, 1, 1)]$. Note that we have $4$ skip connections from the encoder to the decoder. The features from different layers of the encoder are passed through convolution layers with kernel size $3$ and the same number of input/output channels as the input features. The loss function used to train the DPR baseline is:
{\small
\begin{align}
    \mathcal{L}(\theta) = & \Vert I_2 - f_\theta(I_1, L_2)\Vert_1 + \lambda \Vert L_1 - \widehat{L}_1 \Vert + \\ \nonumber
    &\Vert I_1 - f_{\theta}(I_2, L_1)\Vert_1 + \lambda \Vert L_2 - \widehat{L}_2 \Vert_1
\end{align}}%
where $(I_1, I_2)$ are the multi-lit image pair from the multi-lit CelebA dataset, $L_1$ and $L_2$ are the ground truth (GT) lighting vectors for $I_1$ and $I_2$ respectively. Also, $\widehat{L}_1$ and $\widehat{L}_2$ are the estimated source lighting vectors for $I_1$ and $I_2$ images and $f_{\theta}$ denotes the DPR encoder-decoder network parameterized by $\theta$. Note that while training the DPR baseline, we have supervision over the relit image and the estimated source lighting by the network.

\subsection{Details on Pix2Pix baseline}
For the Pix2Pix~\cite{pix2pix2017} baseline, we have a conditional GAN~\cite{cgan} model. The generator consists of a U-net based encoder-decoder. The encoder takes as input the image and the target lighting vector. To concatenate these two before feeding it to the encoder, we first repeat the lighting vector to have the same width and height as the input image. Then, we concatenate this repeated vector with the input image along the channel dimension. The encoder consists of $4$ convolution layers with output channel sizes of $32$, $64$, $128$ and $256$ with ReLU activations and batchnorm, kernel size $5$ and stride length $2$. The decoder also have $4$ deconvolution blocks to upsample the lower resolution feature from the encoder. Also, there are skip-connections between respective layers of the encoder and decoder to allow the flow of high frequency features directly from the input to the output. The discriminator consists of $3$ convolution layers, with $32$, $64$ and $128$ channels and a kernel size of $5$ and stride $2$. The output of the cascaded convolution blocks is then passed through an adaptive average pooling layer and two fully connected layers with $128$ and $1$ output sizes. To have the final score from the discriminator between $0$ and $1$, we pass the final output through a sigmoid. The discriminator takes as input the relit image along the target lighting. We use the following loss to train the Pix2Pix baseline:
\begin{align}
    \label{eq:loss_pix2pix}
    \mathcal{L}(\varphi, \phi) = & \lambda \Vert \mathcal{G}_{\varphi}(I_s, L_t) - I_t \Vert_1 + \\ \nonumber &\log(\mathcal{D}_{\phi}(I_t)) + \log(1-\mathcal{D}_{\phi}(\mathcal{G}_{\varphi}(I_s, L_t))) 
\end{align}
\begin{align}
    \varphi^*, \phi^* = \arg \min_\varphi \max_\phi \mathcal{L}(\varphi, \phi)
    \label{eq:loss_pix2pix2}
\end{align}
where $\mathcal{G}_{\varphi}$ and ${D}_{\phi}$ denote the generator and the discriminator parameterized by $\varphi$ and $\phi$. $I_s$ and $I_t$ and the source and target images and $L_t$ denotes the target lighting. Note that in~\eqref{eq:loss_pix2pix} the first term enforces the relit image to be close to the ground truth target image. We train Pix2Pix baseline on multi-lit CelebA~\cite{liu2015faceattributes} dataset which contains lighting vectors from the SfSNet dataset~\cite{sfsnetSengupta18}. 

\begin{figure}
    \centering
    \includegraphics[width=1  \linewidth]{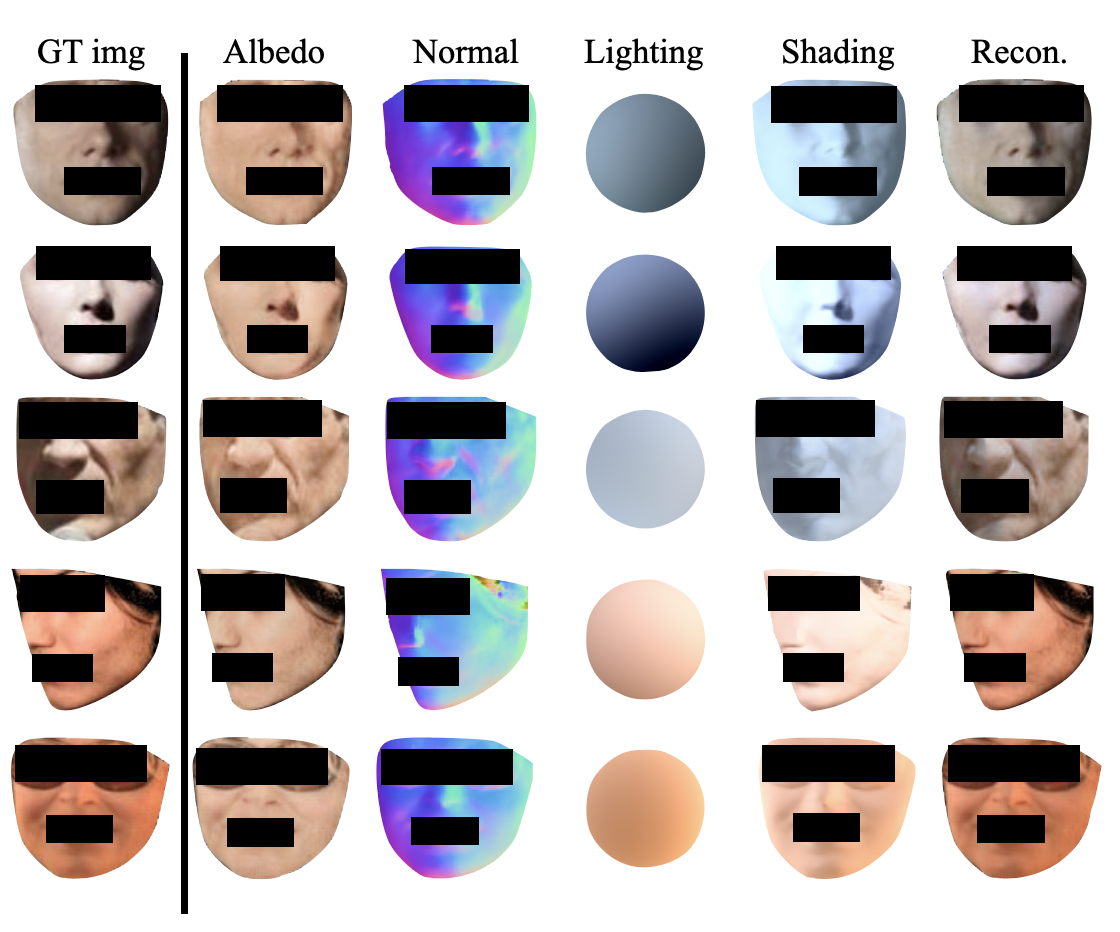}
    \caption{Some challenging examples for the weakly supervised model.}
    \label{fig:fail_examples}
\end{figure}

\begin{figure}
    \centering
    \includegraphics[width=1  \linewidth]{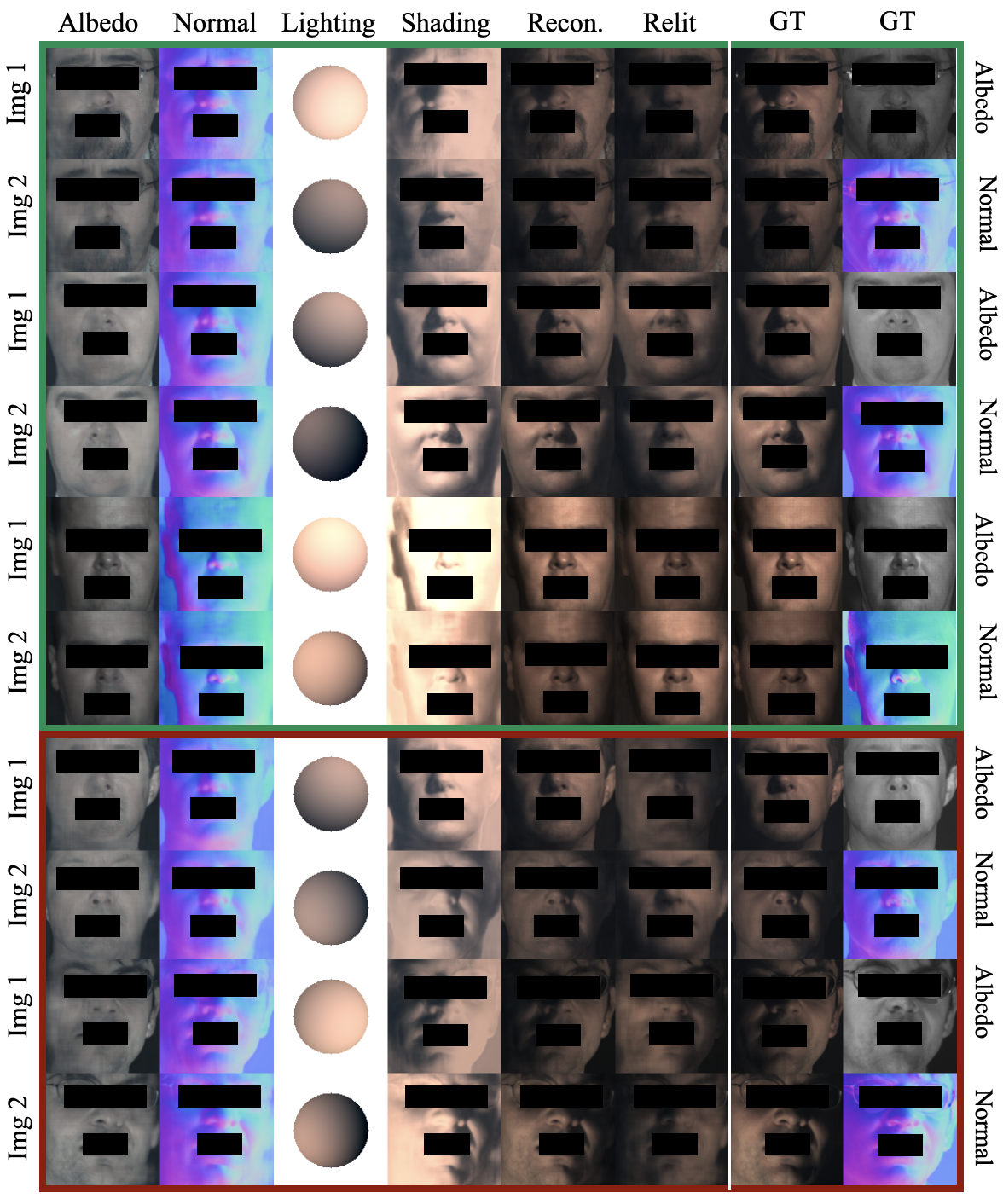}
    \caption{Decomposition and relighting results of the model trained from scratch on the Photoface dataset~\cite{photoface}.}
    \label{fig:photoface_scratch}
\end{figure}

\subsection{Challenging examples}
In Fig.~\ref{fig:fail_examples} we present some challenging examples. Note that, the model has difficulties in the decomposition task for some non-frontal poses. Also, cast shadows alongside partial occlusions result in less accurate decomposition (although still reasonable).

\subsection{Visual results on Photoface}
In Fig.~\ref{fig:photoface_scratch}, we visualize decomposition and relighting results on the Photoface dataset~\cite{photoface} for a model trained from scratch on the Photoface dataset. We provide both successful and failed examples, outlined in green and red. For low-light images or parts of the face under shadow, the recovered albedo has low quality. This then leaks to the relit image, leading to a lower quality relit image and visible artifacts.

\subsection{Decomposition results}
\begin{figure*}
\centering
\begin{minipage}{0.45\textwidth}
\centering
\includegraphics[width=1\linewidth]{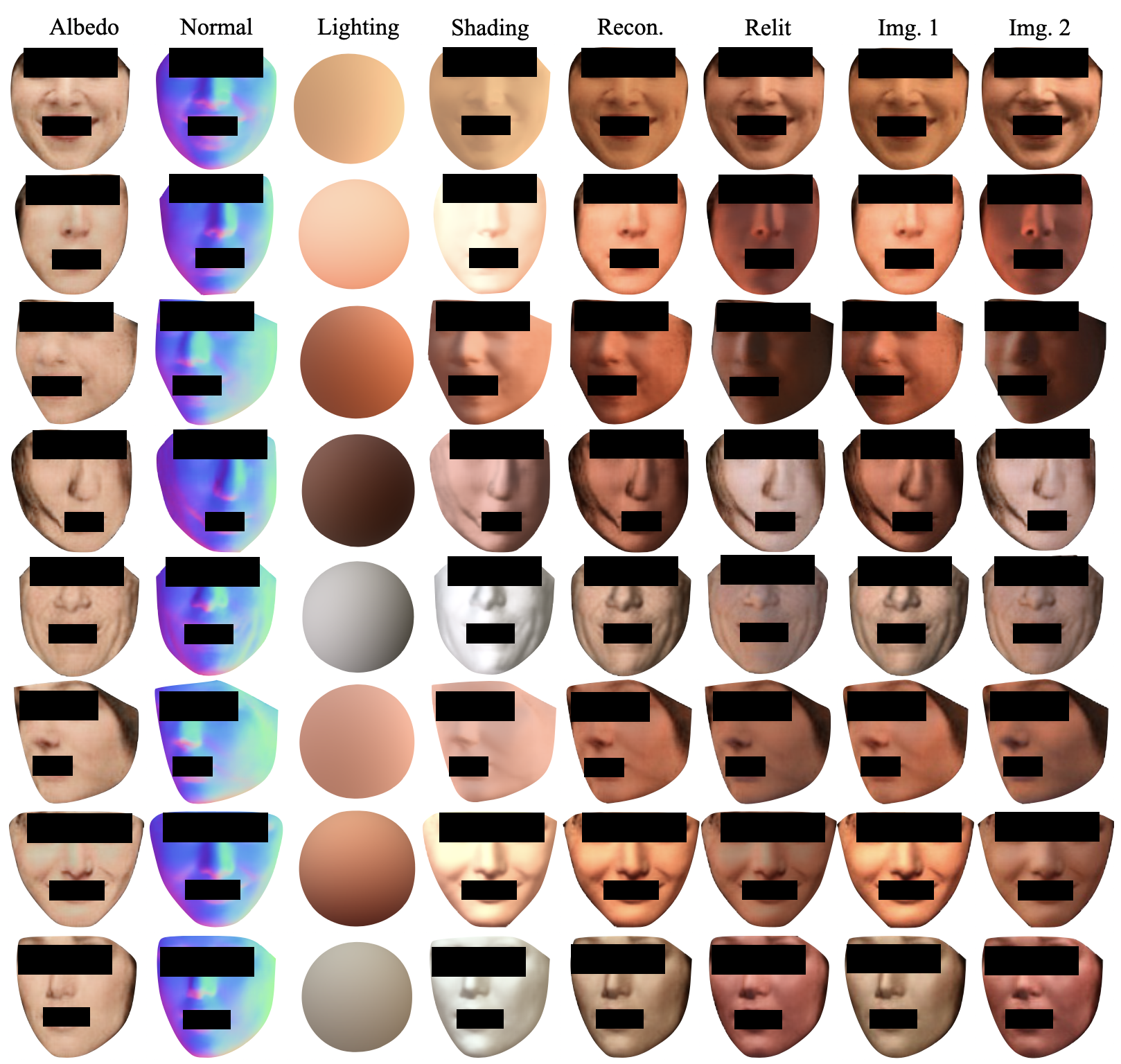}
\vspace{-5pt}
\subcaption{The weakly supervised model. During training we have used light supervision, reconstruction and cross-relighting loss terms.}
\end{minipage}
\hspace{2pt}
\begin{minipage}{0.45\textwidth}
\centering     
\includegraphics[width=1\linewidth]{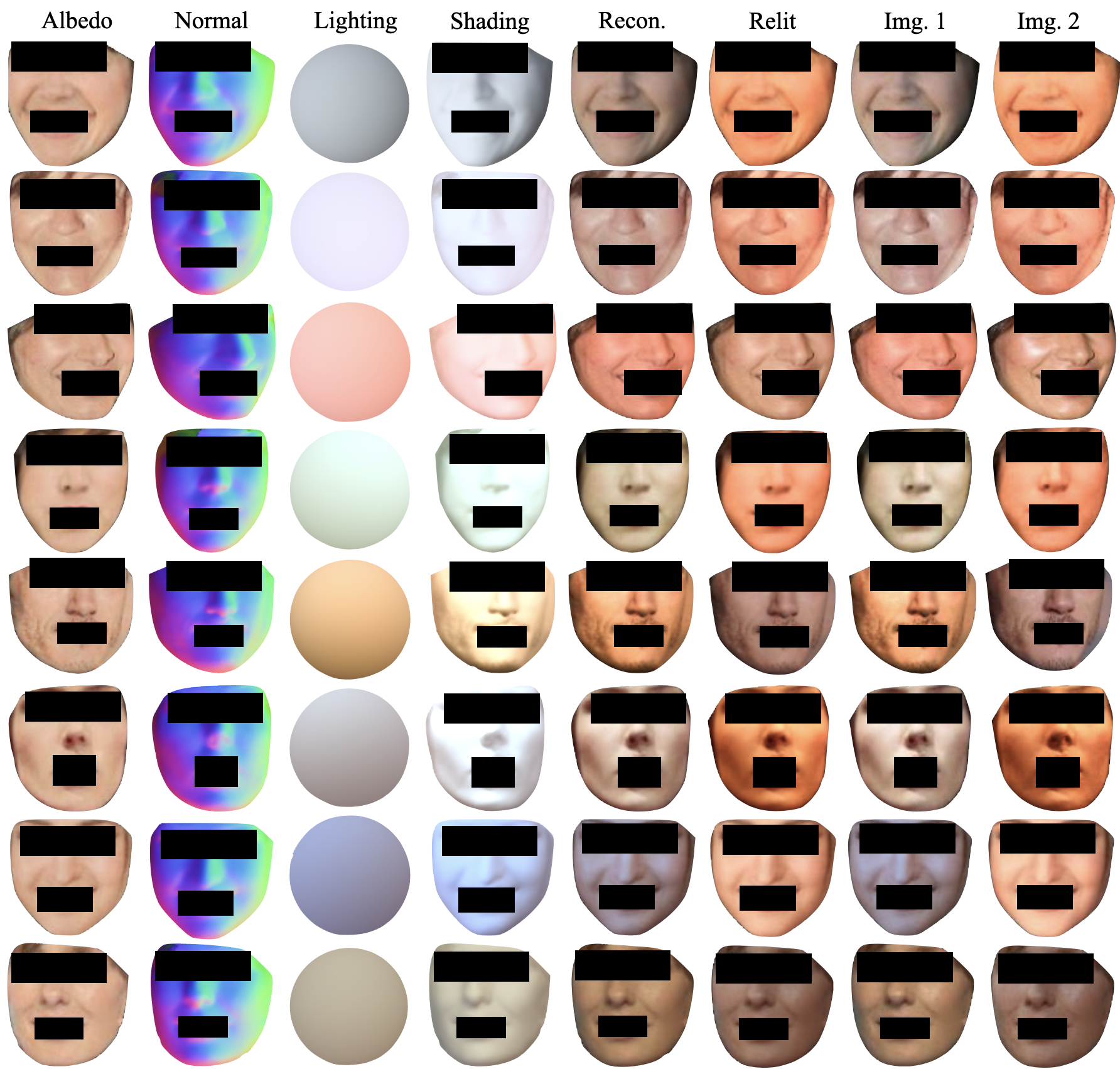}
\vspace{-5pt}
\subcaption{Model trained on a  mixture of synthetic and real datasets. We have full supervision on synthetic samples, while only use $\mathcal{L}_{\textrm{rec}}$ and $\mathcal{L}_{\textrm{relit}}$ for samples from multi-lit CelebA.}
\end{minipage}
\caption{Decomposition results of different models.}
\label{fig:decomp_results}
\vspace{-10pt}
\end{figure*}

In Fig.~\ref{fig:decomp_results}, we present more decomposition and relighting results for the weakly supervised model and the model trained on a combination of synthetic and real datasets. 

\subsection{Visual comparisons to DPR and Pix2Pix}
\begin{figure}
    \centering
    \includegraphics[width=1  \linewidth]{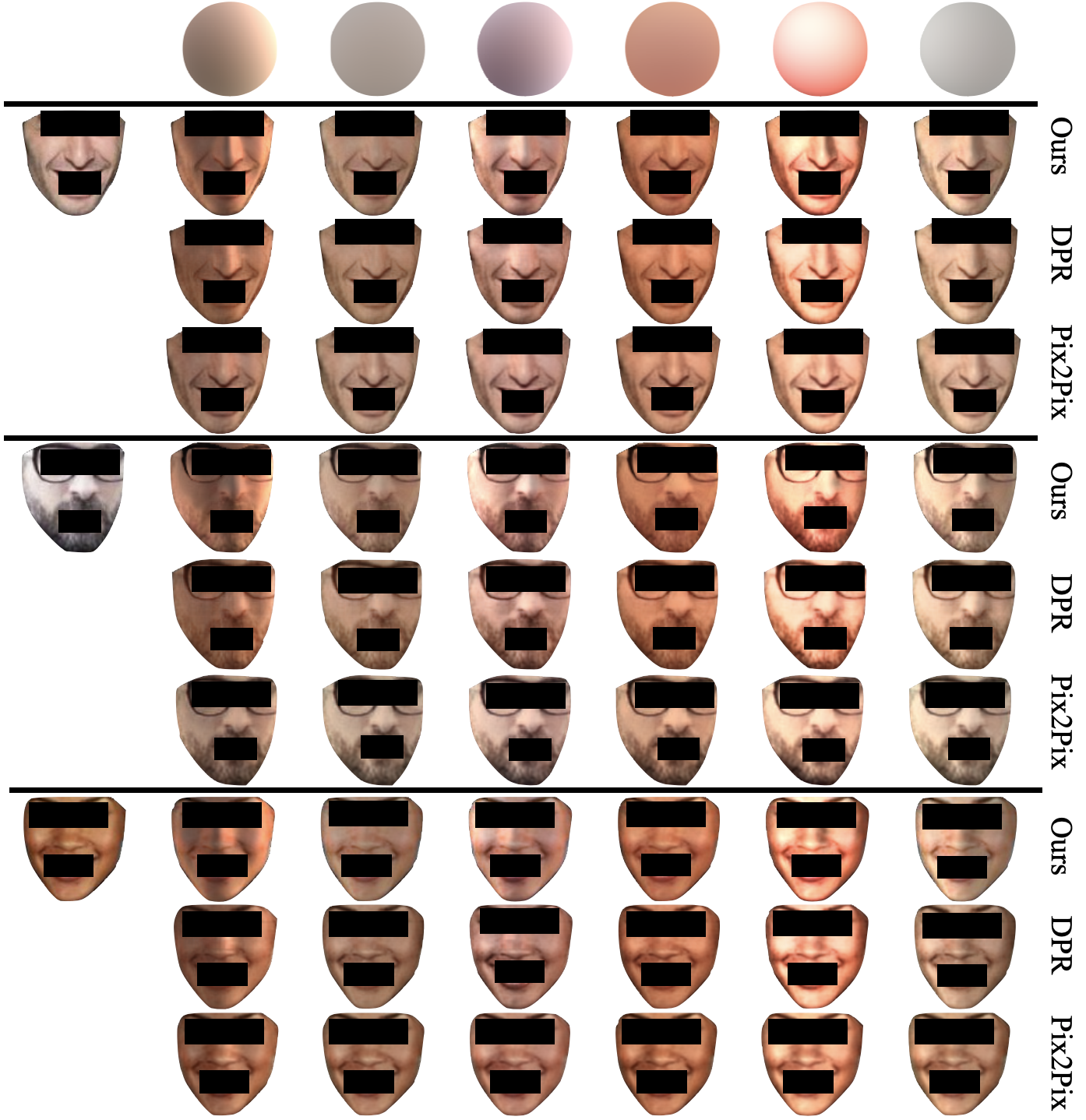}
    \caption{Comparison between the relighting results of our model vs DPR~\cite{DPR} and Pix2Pix~\cite{pix2pix2017} baselines trained on multi-lit CelebA dataset. The target lights are depicted in the first row.}
    \label{fig:relight_dpr4}
\end{figure}

\begin{figure}
    \centering
    \includegraphics[width=1  \linewidth]{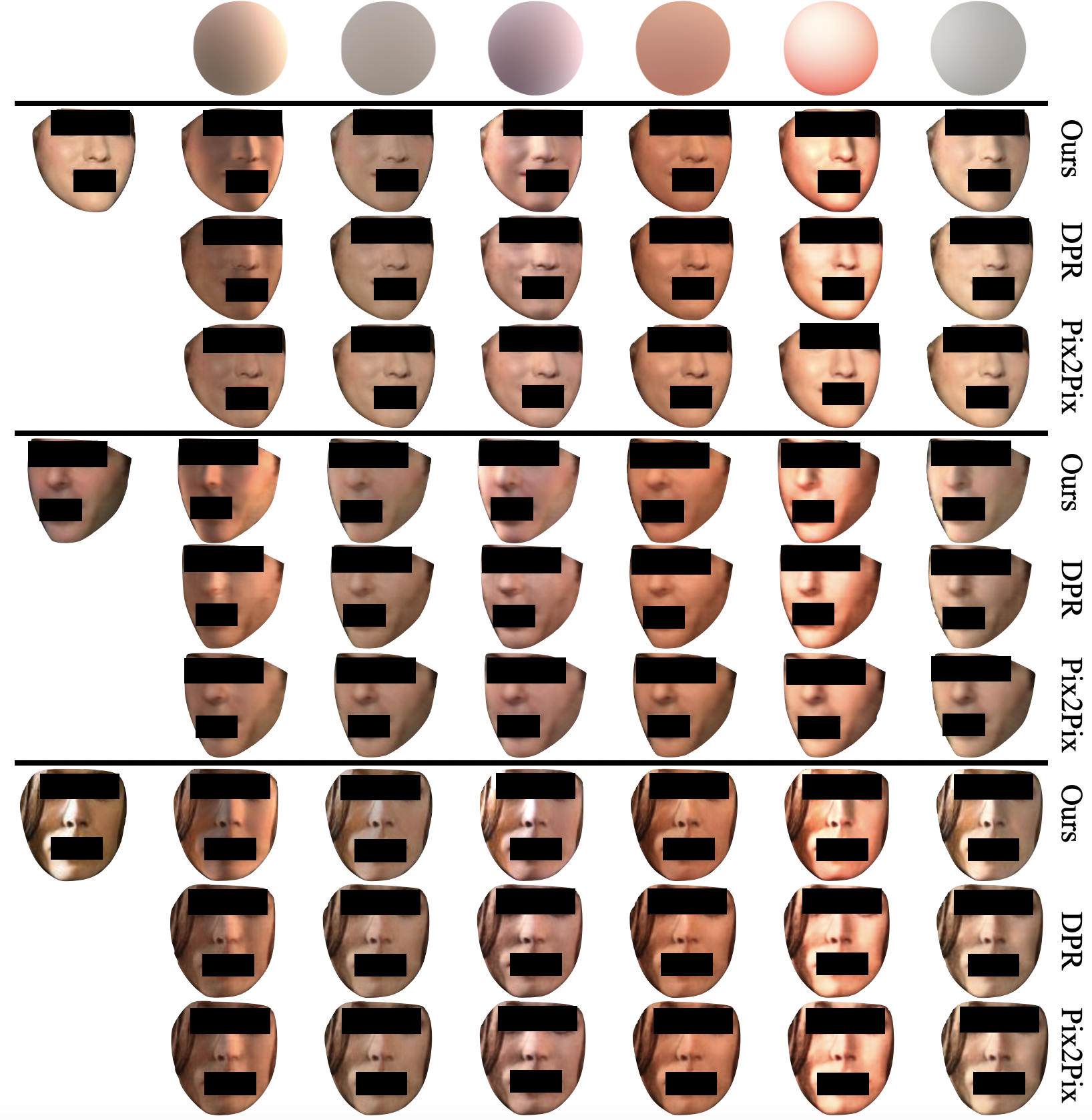}
    \caption{Comparison between the relighting results of our model vs DPR~\cite{DPR} and Pix2Pix~\cite{pix2pix2017} baselines trained on multi-lit CelebA dataset. The target lights are depicted in the first row.}
    \label{fig:relight_dpr5}
\end{figure}

In Fig.~\ref{fig:relight_dpr4}-\ref{fig:relight_dpr5}, we visually compare the relighting performance of our model against the DPR and Pix2Pix models we trained on the multi-lit CelebA dataset. Note that our model compared to the baselines generate relit images that better match the target lighting and look natural.

\subsection{Light transfer}
\begin{figure}
    \centering
    \includegraphics[width=1  \linewidth]{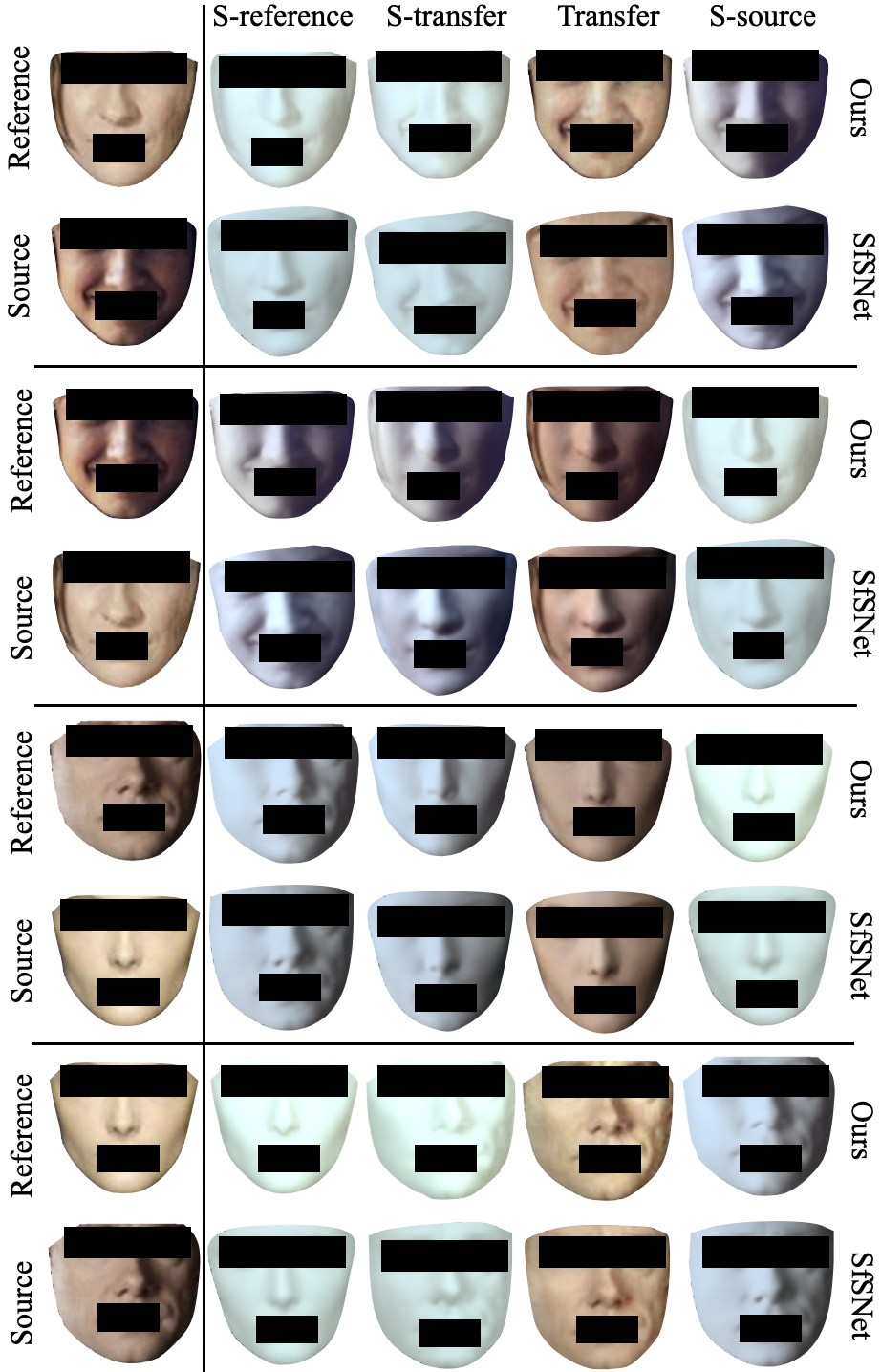}
    \caption{Comparison between our model and SfSNet~\cite{sfsnetSengupta18} for the light transfer task. S-* denotes shading.}
    \label{fig:light_transfer3}
\end{figure}

\begin{figure}
    \centering
    \includegraphics[width=1  \linewidth]{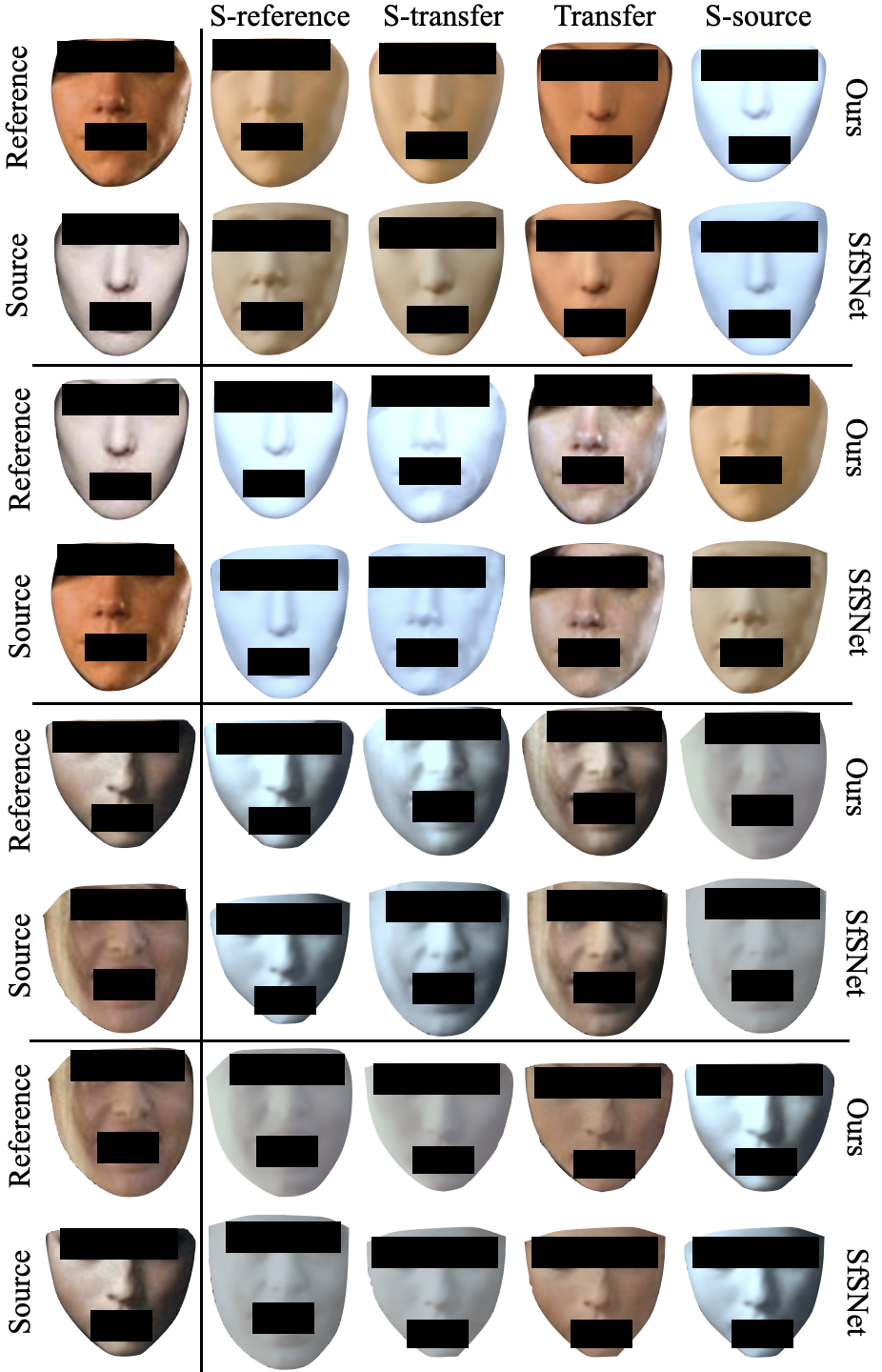}
    \caption{Comparison between our model and SfSNet~\cite{sfsnetSengupta18} for the light transfer task. S-* denotes shading.}
    \label{fig:light_transfer4}
\end{figure}

In Fig.~\ref{fig:light_transfer3}-\ref{fig:light_transfer4}, we present additional visual results and comparisons with SfSNet~\cite{sfsnetSengupta18} on the light transfer task.

\end{document}